# How Linguistics Learned to Stop Worrying and Love the Language Models

Richard Futrell and Kyle Mahowald[1]

Richard Futrell
University of California Irvine, USA; rfutrell@uci.edu

Richard Futrell is Associate Professor of Language Science at the University of California, Irvine. His research integrates information theory, computational linguistics, and psycholinguistics to model how efficiency and memory shape language, earning best-paper honors at ACL and CogSci. He received a Ph.D. in Cognitive Science from MIT and an MA and BA in Linguistics from Stanford.

Kyle Mahowald
The University of Texas at Austin, USA; kyle@utexas.edu

Kyle Mahowald is Assistant Professor of Linguistics at the University of Texas at Austin and works at the intersection of linguistics, cognitive science--, and natural language processing. His papers have been recognized with Best or Outstanding Paper Awards at the computational linguistics conferences EMNLP and ACL. He is also the recipient of a National Science Foundation (NSF) CAREER award. He received his M.Phil in General Linguistics from Oxford in 2011 and his Ph.D. in Cognitive Science from MIT in 2016; his postdoctoral training was at Stanford in Natural Language Processing.

**Abstract**

Language models can produce fluent, grammatical text. Nonetheless, some maintain that language models don't really learn language and also that, even if they did, that would not be informative for the study of human learning and processing. On the other side, there have been claims that the success of LMs obviates the need for studying linguistic theory and structure. We argue that both extremes are wrong. LMs can contribute to fundamental questions about linguistic structure, language processing, and learning. They force us to rethink arguments and ways of thinking that have been foundational in linguistics. While they do not replace linguistic structure and theory, they serve as model systems and working proofs of concept for gradient, usage-based approaches to language. We offer an optimistic take on the relationship between language models and linguistics.

## 1 Introduction

It's 1968, and Norm and Claudette are having lunch. Norm is explaining his position that all human languages share deep underlying structure and has worked out careful theories showing how the surface forms of language can be derived from these underlying principles. Claudette, whose favorite movie is the recently released *2001: A Space Odyssey* and who particularly loves the HAL character, wants to make machines that could talk with us in any human language. Claudette asks Norm whether Norm thinks his theories could be useful for building such a system. Norm says he is interested in human language and the human mind, found HAL creepy, and isn't sure why Claudette is so interested in building chatbots or what good would come of that. Nonetheless, they both agree that it seems likely that, if Norm's theories are right (and he sure thinks they are!), they could be used to work out the fundamental rules and operations underlying human language in general—and that should, in principle, prove useful for building Claudette's linguistic machines. Claudette is very open to this possibility: all she wants is a machine that talks and understands. She doesn't really care how it happens. Norm and Claudette have very different goals, but they enjoy their conversations and are optimistic that they can both help each other.

[1] Author order determined alphabetically by last name.

Fast forward to 2025. Norm has worked for decades on a variety of diverse languages, developing sophisticated theories of linguistic structure. Claudette got more and more interested in engineering, amassing huge amounts of data, and training statistical models. Norm and Claudette slowly fell out of touch over the years. They are planning a reunion lunch: what should they talk about?

Just like Claudette always dreamed of, she now has machines that can talk with her in English, producing grammatical and seemingly sensible utterances. How relevant is it that the architecture of Claudette's machines seems to have nothing to do with the structure of language as identified by Norm and other linguists and cognitive scientists? Claudette might not care: she just wanted a system that worked, and this stuff works. On the other hand, what if Norm was right about the nature of language—does that mean the machines aren't actually as impressive as Claudette thinks, because they are relying on shallow pattern matching? Or are Claudette's machines evidence that Norm's theories were wrong? More generally, what insights do these machines have for the scientific study of language?

One approach Norm could take would be to dismiss Claudette's machines out of hand as irrelevant to human language, as they are not realistic models of the human brain, nor are they intended to be. Under this view, neural networks are no more relevant to linguistics than submarine engineering is to an ichthyologist—just because both submarines and fish can move underwater does not mean that you can learn much about one from studying the other.

This negative view takes several forms. One argument is that Language Models (yes, Claudette's machines are Language Models; henceforth LMs) are so fundamentally different in their architecture from people, and have access to so much more data, that whatever they are doing is so different as to be totally irrelevant for humans (Chomsky et al., 2023; Fox and Katzir, 2024; Bolhuis et al., 2024). Some have denied that language models have actually learned, or ever could learn, the putatively key properties of human language, and thus are either irrelevant to language science or do not challenge any existing ways of thinking (Lan et al., 2024; Fox and Katzir, 2024). A different and more subtle argument is that neural network sequence models could learn to approximate *anything*, so the fact that they seem to learn language (to the extent that they do) is uninformative (Rawski and Baumont, 2023; Moro et al., 2023; Chomsky, 2023; Chomsky et al., 2023; Collins, 2024; Bolhuis et al., 2024). Under this view, LM architectures are like epicycles, the computational technique used by Ptolemy to predict the motions of the planets and the sun in a model that placed the Earth at the center of the universe (de Santillana, 1955; Flynn, 2013). The problem with epicycles is that they can approximate any trajectory arbitrarily well (at the cost of great complexity), so the mere fact that they could be used to capture the motions of the planets with high accuracy tells us nothing about the real nature of that motion. Similarly, if neural networks are just general pattern approximators, then the fact that they can approximate language doesn't tell us anything about language.

We believe these negative views are short-sighted: language models *do* learn non-trivial aspects of linguistic structure, and they do give important insights that change how we should think about language. As language scientists, we ignore them at our peril.

An opposite approach is to dismiss traditional theories of linguistic structure, on the grounds that they turned out to be either useless or of negative value in developing the only known systems that can actually use language as humans do (Jelinek, 2004; Piantadosi, 2023). Under this view, decades of work on the structure of language turned out to be barking up the wrong tree, and the theory of language should be rebuilt on top of the only computational foundation that has been demonstrated to be able to produce and comprehend language to a human level: large neural networks. In our experience, this view is widespread in some engineering and application-focused communities.

This view is also short-sighted. It throws out hard-won analytical discoveries about the structure of language and leaves us adrift in terms of a scientific theory of language, without a clear way to approach the question of *why* human language is the way it is, or even a sense of what the interesting questions are. Moreover, language models are currently most successful in English and other languages with internet-scale data (Blasi et al., 2022). A more complete approach to the science of language will draw on the expertise of documentary linguists, sociolinguists, anthropologists, and community stakeholders, and it will integrate the insights from decades of linguistic inquiry.

Thus, we advocate a third view (joining fellow travelers in linguistics, cognitive science, and philosophy who have advocated for some form of this middle ground, e.g., Smolensky, 1988; Pater, 2019; Portelance and Jasbi, 2023; McGrath et al., 2024; Millière, 2024; Potts, 2025; Chesi, 2025): language models are not a complete theory of language—in fact, no one has such a theory—but they are hugely informative about language and its structure, learning, processing, and relationship with the larger structure of the mind. Language models have set off an intellectual explosion in cognitive science, machine learning, philosophy of mind, and other fields, in which longstanding ideas have been overturned; novel ideas are emerging; and disciplinary boundaries are dissolving. Linguistics has a chance to stand at the center of this huge intellectual ferment, and would be remiss to isolate itself intellectually on the basis that language models don't look like existing theory. Language science can contribute (and in fact already has contributed) to the development of language models, and language models can contribute (and in fact already have contributed) insights about language. Norm and Claudette have a lot to talk about.

# 2 Statistical models of language have outperformed expectations

## 2.1 A brief history of statistical language learning

The generative school of modern linguistics arose from an argument against the idea that the structure of language could be learned from the statistics of language data. Chomsky perhaps most famously illustrated this by contrasting (1) "Colorless green ideas sleep furiously" with "(2) Furiously sleep ideas green colorless." Even though neither sentence makes semantic sense, (1) is clearly grammatical in a way that (2) isn't. As Chomsky (1957) put it: "The notion 'grammatical in English' cannot be identified in any way with the notion 'high order of statistical approximation to English.' It is fair to assume that neither sentence (1) nor (2) (nor indeed any part of these sentences) has ever occurred in an English discourse. Hence, in any statistical model for grammaticalness, these sentences will be ruled out on identical grounds as equally 'remote' from English."

The effective conclusion from these arguments was that linguistic structure could only be characterized in terms of formal systems, based on rules or constraints and operating over structured arrays of symbols (Chomsky, 1965). Early on, the expectation was that such systems would form the basis for language technologies such as machine translation and question answering systems (Hays, 1960; Winograd, 1972; Hutchins, 1981). It was believed that these formal systems should be constructed by linguists since the task of learning them from data was hard or impossible. Yet despite myriad efforts to build machine translation systems, grammatical parsers, and other tools, linguistic competence remained elusive for machines. Symbolic approaches that sought to elucidate rules and structures often proved unable to capture all the exceptions and complexity that characterized natural language as it is used.

By the late 1980s and 1990s, statistical learning had a major renaissance in natural language processing (Brown et al., 1990; Manning and Schütze, 1999; Pereira, 2000), and started to reappear in the human language learning literature as well (Saffran et al., 1996). Yet despite success at various natural language processing tasks, these techniques seemed frustratingly unable to get past approaches that simply counted up words and phrases (Wang and Manning, 2012; Arora et al., 2017). The connectionist movement in the 1980s and 90s seemed promising (Rumelhart and McClelland, 1986; Smolensky, 1988; Elman, 1990a) (and turned out to be the most important forerunner of modern language models), but there were well-justified concerns about the ability of these approaches to scale up and to represent the rules and structures that seem to characterize language (Pinker and Prince, 1988), despite accounts of how connectionist models could in principle implement rule-like symbolic behavior (Smolensky, 1990).

Even researchers optimistic about the role of statistical learning in language acquisition were, in the recent past, deeply skeptical that end-to-end neural approaches would succeed on interesting linguistic tasks. Representatively, Chater et al. (2006) wrote, "connectionism is no panacea here; indeed, connectionist simulations of language learning typically use small artificial languages, and, despite having considerable psychological interest, they often scale poorly." Tenenbaum et al. (2011) wrote, "connectionist models sidestep these challenges by denying that brains actually encode such rich knowledge, but this runs counter to the strong consensus in cognitive science and artificial intelligence that symbols and structures are essential for thought." Nevertheless, given the way that linguistics and machine learning developed and evolved, it now sometimes seems to be taken for granted in practice that ideas from linguistic theory will not form the basis of proficient language processing systems.

Throughout the statistical renaissance, generative linguists continued to claim that statistical methods would never solve interesting problems related to learning linguistic structure. In a review article, Everaert et al. (2015) reiterated the claim that statistical approaches would be fundamentally unable to distinguish sentences like "How many cars did they ask if the mechanics fixed?" from ungrammatical ones like "*How many mechanics did they ask if fixed the cars?", or that they would only be able to do so if provided with certain built-in formal structures. Berwick et al. (2011) were skeptical that recurrent neural networks could ever be much more powerful than bigram models.

Against this backdrop, the fortunes of connectionist models started changing in the 2010s—as a result of new techniques and, most importantly, due to increased computational power that made training neural models much more efficient (Hinton et al., 2006). By 2016, neural models showed rudiments of grammatical generalizations like subject–verb agreement (Linzen et al., 2016). Over the coming years, the success of models at acquiring linguistic abilities continued to grow (Futrell et al., 2019a; Wilcox et al., 2018; Manning et al., 2020; Hu et al., 2020; Warstadt and Bowman, 2022; Mahowald et al., 2024).

The growth from early neural models in, e.g., 2011 to now is remarkable from a historical perspective, and was surprising to virtually everyone in the field at the time. Sutskever et al. (2011) introduced an at-the-time state-of-the-art recurrent neural network that produced output like "In the show's agreement unanimously resurfaced. The wild pasteured with consistent street forests were incorporated by the 15th century BE." and praised it as a "surprisingly good language model," noting the "richness of their vocabularies," that the "text is mostly grammatical," and "parentheses are usually balanced over many characters." Compared to today's language models, the consistent street forests seem a long way away: 2011 might as well have been the 15th century BE. Today's language models (e.g., ChatGPT or Claude) produce seemingly endless streams of highly fluent and grammatical (if not always perfectly humanlike) text and form the basis of the most successful natural language processing approaches.

## 2.2 Neural LMs learn nontrivial linguistic structure

The rapid development of these capacities has been treated with healthy skepticism. LMs have access to vast amounts of data and could be simply memorizing or relying on cheap heuristics — giving the appearance of linguistic competence without really having it.

The most naïve way that one might think to test for grammatical competence in language models would be to see whether they assign higher probability to grammatical strings than ungrammatical ones. However, this would not work, even for an ideal statistical model, for reasons related to the point that Chomsky (1957) was making. For example, the ungrammatical string "Snails died the old" has a probability of $\sim 2^{-49}$ under the GPT-2 language model, whereas the grammatical string "The ancient crustaceans expired" has a lower probability of $\sim 2^{-55}$, simply because it uses lower-frequency words (Wilcox et al., 2023a).

But the failure of this naïve comparison does not mean that language models lack grammatical knowledge. In general—for language models and in real language use— many factors influence the probability of a string, of which the grammar of the language (as represented in the language model or in the mind of a human speaker) is only one: word frequency, utterance length, online processing constraints such as memory limitations, and plausibility given world knowledge all feed into the probability of an utterance. Indeed, not only utterance probabilities, but also comprehension accuracy, reaction time, and indeed any psychometric dependent variable are affected by all of these factors jointly—including the subjective grammaticality judgments that form the basis of much of the formal syntax literature (Kluender and Kutas, 1993; Hofmeister et al., 2013; Mahowald et al., 2016; Lau et al., 2017). Therefore, if we want to look for evidence that language models represent linguistic structure, we need to search more carefully.[2]

The key is to isolate linguistic structure from these other factors through controlled experimental studies and through probing LMs' internal states. Experimentally, from sufficient performance data, one may infer an underlying formal cognitive structure, no matter whether the implementation substrate is a brain or a neural network (Piantadosi and Gallistel, 2024). This is the standard procedure in linguistics, where data consisting primarily of acceptability judgments is used to postulate underlying linguistic competence. This approach can be applied just as well to LMs.

One form of such a study is to conduct behavioral comparisons of minimally different sentences. For example, LMs may be fed sentences such as "The keys to the cabinet are on the table" and the ungrammatical "The keys to the cabinet *is* on the table", and the conditional probabilities assigned to the verb form "are" versus "is" are compared. If the model understands how agreement works in English, then the probability for "are" should be higher than "is". This comparison controls for plausibility, since the context is the same for the two alternatives. The task could not be solved by any $n$-gram model with $n < 5$ (and higher-order $n$-gram models can be ruled out by adding more words between "keys" and "is/are" in the stimuli). The lexical frequency of "is" versus "are" can be controlled through a more elaborate experimental design, with four conditions in a $2 \times 2$ design crossing the grammatical number of the subject with the grammatical form of the verb (as done by Marvin and Linzen, 2018). These are usual procedures in psycholinguistics, where experimenters develop controlled sets of sentences and then measure dependent variables like reading times. The same methodology can be applied to language models with probability as the dependent variable, aiming to ascertain whether models are sensitive to linguistic structure (Linzen et al., 2016; Futrell et al., 2019b).

[2] Nor is it necessarily sensible to evaluate models based on simply prompting them and asking whether sentences are grammatical or not (Hu and Levy, 2023) (as done by Dentella et al., 2023). To see this, consider that many linguists have held that grammatical knowledge can be represented by a formal system like a context-free grammar; even if a context-free grammar is a good model of linguistic knowledge, it could not be prompted in natural language to respond about whether a sentence is grammatical or not. Grammatical knowledge does not imply that a system can make explicit grammaticality judgments when prompted in natural language.

Such studies have revealed behavioral patterns consistent with neural networks representing formal linguistic structures, in cases such as subject–verb agreement (Linzen et al., 2016; Bernardy and Lappin, 2017; Gulordava et al., 2018), filler–gap dependencies (Wilcox et al., 2018, 2023a; Kobzeva et al., 2023; Suijkerbuijk et al., 2023), and recursive embedding of clauses (Futrell et al., 2019b; Wilcox et al., 2019a; Hu et al., 2020), all of which involve highly nontrivial formal structures which statistical models failed to capture in previous work. Example results for subject–verb agreement from GPT-2 are shown in Figure 1: we see that grammatical verb forms are relatively more probable than matched ungrammatical verb forms in all but a few cases (human accuracy in producing the right verb forms in such sentences is 85%; Marvin and Linzen, 2018).[3] The results indicate that the model can represent the non-local structural dependency between the subject of the sentence and the matrix verb.

Of course, this work has not all been positive in its results. Models can be "right for the wrong reasons" (McCoy et al., 2019), adopting shallow heuristics which make correct predictions on certain test sets but do not generalize properly, or models may show mastery of linguistic form without a concomitant ability to understand the implications of utterances (Weissweiler et al., 2022)—in effect mastering linguistic form without mastering linguistic function (Mahowald et al., 2024). As an example of how shallow heuristics may lead to the illusion of good performance, a simple *n*-gram model performs fairly well on some subsets of paired grammatical and ungrammatical sentences of the BLiMP dataset (Warstadt et al., 2020) (although not as well as a neural LM), which does not explicitly control for *n*-gram frequency (Vázquez Martínez et al., 2023).[4]

The possibility that models generalize on the basis of shallow heuristics motivates deeper investigation. We discuss two approaches here. The first involves controlling a model's training data and then observing its generalizations on evaluation data that is unlike anything in the training data (Jumelet et al., 2021; Feng et al., 2024b; Misra and Mahowald, 2024; Leong and Linzen, 2023; Yao et al., 2025). This approach allows one to study true *generalization* in language models, rather than shallow memorized patterns. Ahuja et al. (2025) trained Transformer language models on a corpus of English-like text that has been constrained so that subjects and verbs are always adjacent. That is, the corpus contains sentences like "I saw the key to the cabinets", but never something like "The key to the cabinets is on the table". The question is, then, if the model trained in this way will generalize in the way that humans do, preferring "The key to the cabinets is on the table" (where the form of the verb depends on the subject of the sentence) over something like "The key to the cabinets *are* on the table" (where the form of the verb depends on the linearly previous verb, which is perfectly consistent with the training data). Ahuja et al. (2025) find that neural language models do make the human-like generalization, providing evidence that they can learn the formal structure underlying nonlocal subject–verb agreement without ever observing nonlocal subject–verb agreement (see also Patil et al., 2024).

[3] It has been objected that, while humans make errors in verb agreement and other syntactic constructions, they have an underlying error-free competence which may be distinguished from their errorful performance (Chomsky, 1965, Ch. 1), whereas the neural networks only have errorful performance with no underlying competence (Fox and Katzir, 2024). We believe this argument is mistaken for two reasons. First, it is indeed possible to define (multiple) competence–performance distinction(s) in neural networks, for example, by separating the knowledge contained in embeddings versus the performance of a linear decoder in extracting that knowledge (Pimentel et al., 2020; White et al., 2021; Csordás et al., 2024). Second, even for humans, we only ever have access to performance data (including subjective grammaticality judgments), and we infer mental constructs such as competence, I-language, etc., on the basis of that data (Piantadosi and Gallistel, 2024). We may make the same inferences based on neural network performance data. Competence is not a datum to be explained, but rather a theoretical construct that is useful for explaining performance data—it is a real pattern (Dennett, 1989; Nefdt, 2023) in the sense we will discuss in Section 4.1.

[4] We note that these kinds of criticisms are harder to apply to studies based on more carefully controlled sets of sentences (for example, Marvin and Linzen, 2018; Futrell et al., 2019b; Hu et al., 2020; Wilcox et al., 2023a).

Another approach is to dig into the language models' internal states. Large language models have a reputation of being black boxes, whose internal processes and representations are inscrutable, but researchers have been gaining growing traction on understanding their inner workin.ss

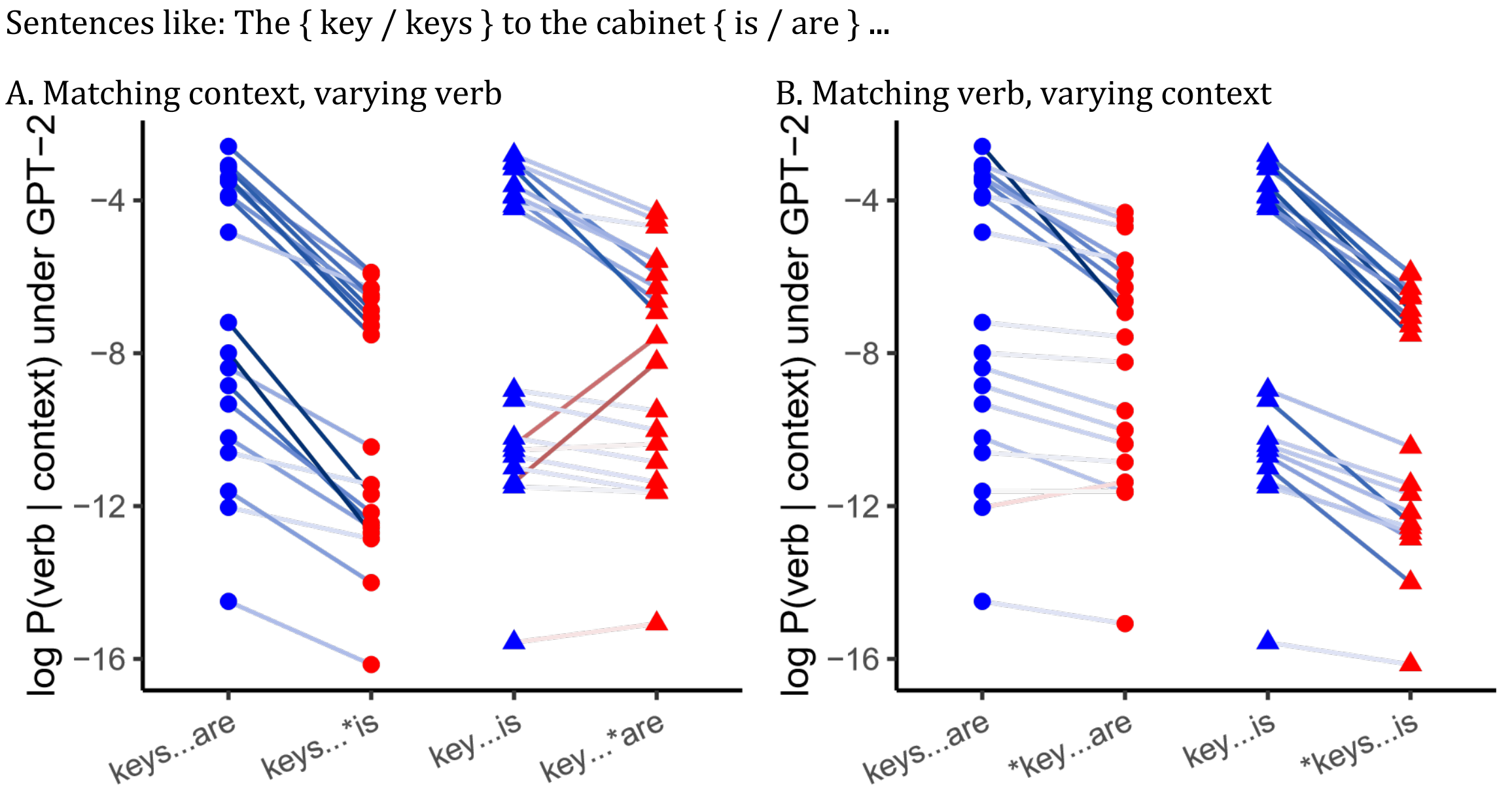

Figure 1: Language models show sensitivity to linguistic structure. Figure shows data from targeted evaluation of subject–verb agreement in GPT-2 from Gauthier et al. (2020), based on design and materials from Marvin and Linzen (2018). Each point represents the log conditional probability for a verb (e.g., singular *is* or plural *are*) in a particular sentence. **A.** Blue points are for grammatical verbs, and red for ungrammatical, in matching contexts. The difference in log probability between grammatical and ungrammatical verbs in identical contexts is indicated by a line. For plural subjects, the grammatical verb is always higher probability. For singular subjects, the grammatical verb is higher probability in 79% of cases. Human accuracy on this task is 85% (Marvin and Linzen, 2018). **B.** Blue points for grammatical verbs, and red points for the same verbs in a context that makes them ungrammatical. The correct context makes the grammatical verb higher probability in all cases for the singular verb *is*, and in 95% of cases for the plural verb *are*.

Approaches include probing, where one attempts to decode linguistic features from the internal representations of neural networks, and causal interventions, where model internals are changed and the resulting output changes are observed (Hewitt and Manning, 2019; Chi et al., 2020; Voita and Titov, 2020; Manning et al., 2020; Papadimitriou et al., 2021; Ravfogel et al., 2021; Lampinen, 2024; Diego-Simón et al., 2024, among others). Such studies have revealed increasing evidence of grammatical structure in models. We consider these kinds of studies, particularly ones that probe models *causally*, to be promising avenues for linguistics and cognitive science (see Section 4.2).

A caveat to these findings is that the various evaluations of structural representation in language models have mostly used English (or a handful of other languages) as the target (Blasi et al., 2022). An important area for future work is to extend the empirical scope of these studies to a wider set of languages (work which is ongoing; see Jumelet et al., 2025, for a multilingual grammatical benchmark).

While there is disagreement about how much language models capture more complex formal patterns (Vázquez Martínez et al., 2023; Lan et al., 2024; Someya et al., 2024) or to what extent they can be said to "understand" (Bender and Koller, 2020) or refer to things in the world (Mandelkern and Linzen, 2024; Lederman and Mahowald, 2024), this is not the right place to adjudicate these debates. What is clear is that language models have learned nontrivial formal linguistic patterns better than many expected was possible. That is to say, they really *have* learned something about linguistic structure, in a way that problematizes earlier claims about the role of statistics in language learning. They are not *purely* relying on heuristics or memorization or cheap tricks. They have, to a meaningful extent, learned "the real thing"—that is, the thing that we care about, as language scientists interested in questions like how languages are learned, how they are processed, how and why they vary, and where they come from.

We next ask why this is interesting for the science of language, and how the science of language productively integrates these insights.

# 3 The success of LMs is interesting for the science of language

> "...nor did Pnin, as a teacher, ever presume to approach the lofty halls of modern scientific linguistics, that ascetic fraternity of phonemes, that temple wherein young people are taught not the language itself, but the method of teaching others to teach that method; which method, like a waterfall splashing from rock to rock, ceases to be a medium of rational navigation but perhaps in some fabulous future may become instrumental in evolving esoteric dialects—Basic Basque and so forth—spoken only by certain elaborate machines."
>
> Vladimir Nabokov
> *Pnin*

LMs learn nontrivial aspects of language. But one could still object that this result is not relevant to language science because human language is the product of the human brain (and/or vocal tract, body, social environment, or what have you), whereas LMs have a totally different architecture and different objective. Perhaps these differences are so great that a language scientist has practically nothing to gain from studying language models, in the same way that there is limited value to an ornithologist studying jet engines, even though both bird wings and jet engines enable flight (Kodner et al., 2023).

Here we argue that the success of language models is relevant, indeed important, to the science of language, because (1) there are good reasons to believe that there should be parallels between engineering artifacts like language models and human language, and (2) the success of language models in learning from text upends ways of thinking that are deeply ingrained in generative linguistics and parts of cognitive science, and (3) language models are already connected to statistical and probabilistic traditions in linguistics, and in fact arose out of them to a large extent. As such, LMs aren't alien invaders into linguistics from engineering. Rather, they are tools similar to those that have long been used to answer fundamentally linguistic questions.

## 3.1 Parallels between engineering models and cognition

From the study of vision, there is strong precedent for the idea that neural networks developed purely for practical applications can tell us a great deal about cognition as it is implemented in the brain. Similarly, there is good reason to think that LMs will be informative for the study of human language. To elucidate this example, we briefly review, in simplified form, the development of the neuroscience of vision and the role of neural networks in it.

Hubel and Wiesel (1959) discovered that early processing of visual information is performed by neurons that are selectively responsive to *edges* in the visual input. After this discovery, the question became why early visual processing works this way. Olshausen and Field (1996), building on information-theory-inspired intuitions from Barlow (1961, 1989), showed that edge detectors resembling those found in the visual cortex were the generic solution to the problem of representing visual information accurately in a neural network under a constraint that only a small number of units should be active (that is, a sparsity constraint, arising ultimately from a power constraint on neural firing). Even better explanation—both in the sense of explanatory depth and predictive accuracy—came from the engineering of artificial visual systems, in particular the development of AlexNet, a large (for the time) hierarchical convolutional neural network for image classification (Krizhevsky et al., 2012). Yamins et al. (2014) showed that this architecture, when trained to do object recognition, not only developed edge filters in its early layers, but also receptive fields in later layers corresponding to later layers of visual processing in the primate brain. The overall picture that emerges from this line of work is: the neuronal organization of visual processing is determined by the function of finding a sparse code to identify and manipulate objects in the environment, a function largely shared between biological and artificial systems.

Here, the artificial system provided guiding insights for understanding the natural system.[5] Something similar is now happening in the science of language processing, where the internal representations developed by language models are predictive of activation patterns in language areas of the brain (Goldstein et al., 2022; Caucheteux et al., 2023; Hosseini et al., 2024b; Rathi et al., 2024), and predictability as estimated by a language model is an important factor in studies that predict neural activity (Stanojevíc et al., 2023; Zhao et al., 2025). The picture is not yet as clear as it is in vision—this is an area of active research, and there are no animal models we can use to get the plentiful high-resolution controlled neural data we would like—but there is precedent to think that the artificial system will be informative about the natural system here too.

Why did this work in vision? To explain the success of the approach, Cao and Yamins (2021) introduce the *Contravariance Principle*: the idea that, if we want to uncover solutions to problems that are common between brains and models, we should focus on hard problems. The reason is that, if a computational problem is hard in the sense that it requires satsifying multiple potentially competing constraints at once, then there are likely to be only a small number of ways to solve it. So we expect different systems that solve the same hard problem (for example, neural networks and the brain) to converge to the same solution (see also Huh et al., 2024; Hosseini et al., 2024a). If a problem is relatively simple, then we might expect many different solutions to work. For instance, there are famously scores of algorithms that have been proposed for sorting numbers in an array.

Vision is not like sorting a list of numbers: it is a hard problem that requires satisfying many constraints (e.g., fast processing, reliable transmission of input, invariance to different light conditions, among many others). Therefore, we can expect that *any* system that solves the problem will share core features with any other. We see this not only in the comparison of humans vs. neural networks, but also in the comparison of humans vs. monkeys (Rajalingham et al., 2018), and primates vs. animals whose visual cortices arise from totally different evolutionary phylogenies, such as cephalopods (Pungor et al., 2023).

[5] There remains controversy over the extent to which deep neural networks are the best model of human vision (see Bowers et al., 2023, for a more skeptical take along with spirited replies), but even skeptics of neural models judge that this research program has been fruitful.

The problem of learning and using language is also not like sorting a list of numbers: an entity that learns and processes language has to satisfy many constraints (storage of lexical items, generalization to novel contexts, fast processing, etc.). And as we have discussed, large Transformer-based models do learn and deploy nontrivial aspects of language. If the Contravariance Principle is right, then it is reasonable to expect some degree of commonality in the solutions that exist in the brain and in Transformers. Indeed, the massive literature on linguistic interpretability in Transformers, which we partially reviewed in Section 2.2, has revealed representational strategies in neural networks that lead to correct novel predictions about human performance (Lakretz et al., 2021), and the way that Transformers process syntactic features like agreement cues has close parallels with independently proposed cognitive frameworks for modeling human language processing based on cue-based retrieval (Lewis and Vasishth, 2005; Ryu and Lewis, 2021; Timkey and Linzen, 2023).

Thus, it is just not the case that these models can be dismissed out of hand because they are different from the brain. To do so would be to miss out on a source of *in silico* insight and inspiration which has been hugely useful in other areas of cognition.

## 3.2 Understanding LM success requires rethinking language learning

### 3.2.1 The significance of the learning problem in linguistics and cognitive science

For a century, cognitive scientists, linguists, computer scientists, and philosophers developed formal theories of how language learning—or any learning—must work. The key concept has been what knowledge is brought to the learning process by a learner beyond the data, known as inductive bias in the machine learning literature (Mitchell, 1980; Goyal and Bengio, 2022), as illustrated in Figure 2. We can think of a learner as a device that takes in some data and outputs a hypothesis, or a set of hypotheses, or a probability distribution over hypotheses, for the underlying process generating the data. For example, given a bunch of data points in a 2D plane, one hypothesis would be that the points are generated along a line in the plane; this is the assumption underlying linear regression. Given linguistic utterances as data, one might hypothesize a certain grammar underlying the data.

The key issue is that, in a very general sense, the data fundamentally underdetermine the hypothesis that the learner arrives at, because any data is compatible with many, many hypotheses. In Figure 2A, where the learning data is points and the hypotheses are lines, both the straight and the curvy lines fit the learning data equally well; the fact that the straight line seems like the right hypothesis—and is likely the one that humans would seize on—therefore cannot be a function of fit to the data. It must be a function of fit to the data plus something that biases a learner to favor one hypothesis over another, even among hypotheses that fit the data equally well: this is inductive bias. Inductive bias is something that the model (or modeler) brings to the problem, not something inherent in the data.[6]

A. Numerical data

[6] We use the term inductive bias here to refer to anything that causes a learner to prefer one hypothesis over another, beyond any function of input data (Mitchell, 1980). Such biases exist and are necessary even in learning algorithms that are not strictly 'inductive', and they may or may not be identifiable as Bayesian priors.

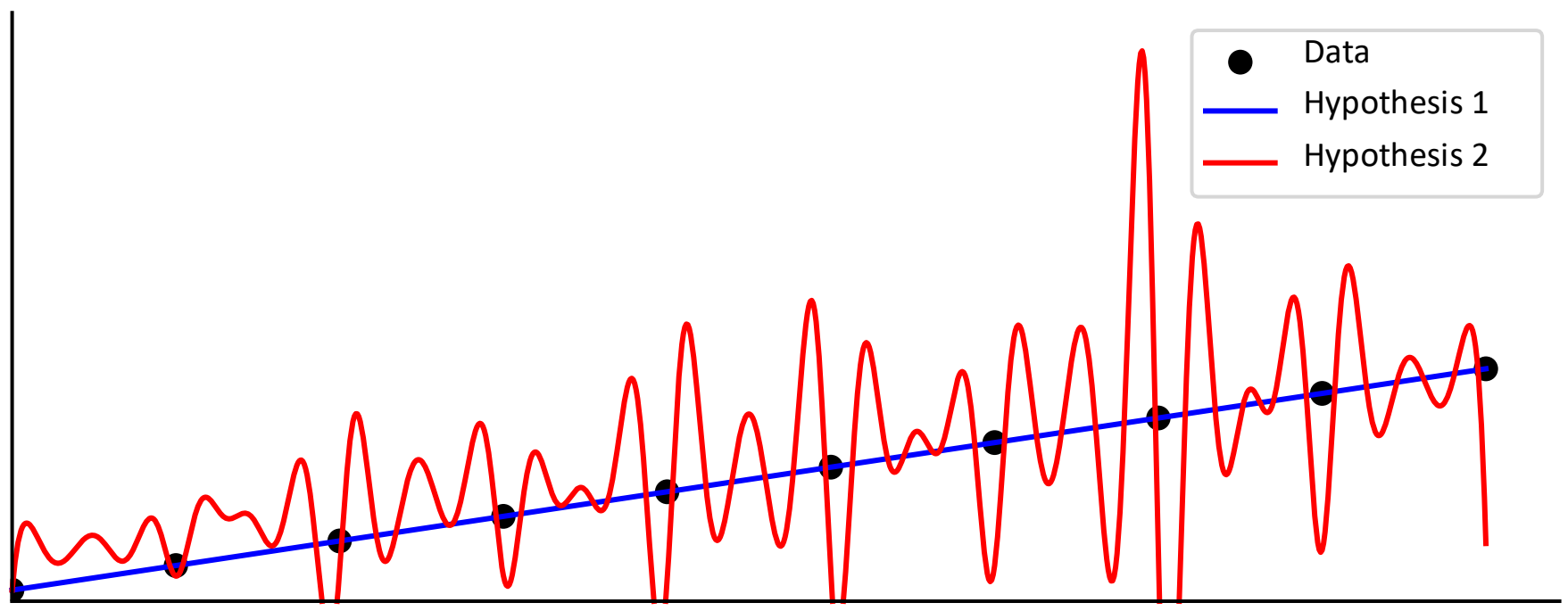

B. Linguistic data

| Data | |
|---|---|
| The key is pretty. | The keys are pretty. |
| Are the keys pretty? | I like the keys to the cabinet. |
| The man is happy. | Is the man happy? |
| The man who is tall is happy. | I like the man who is tall. |

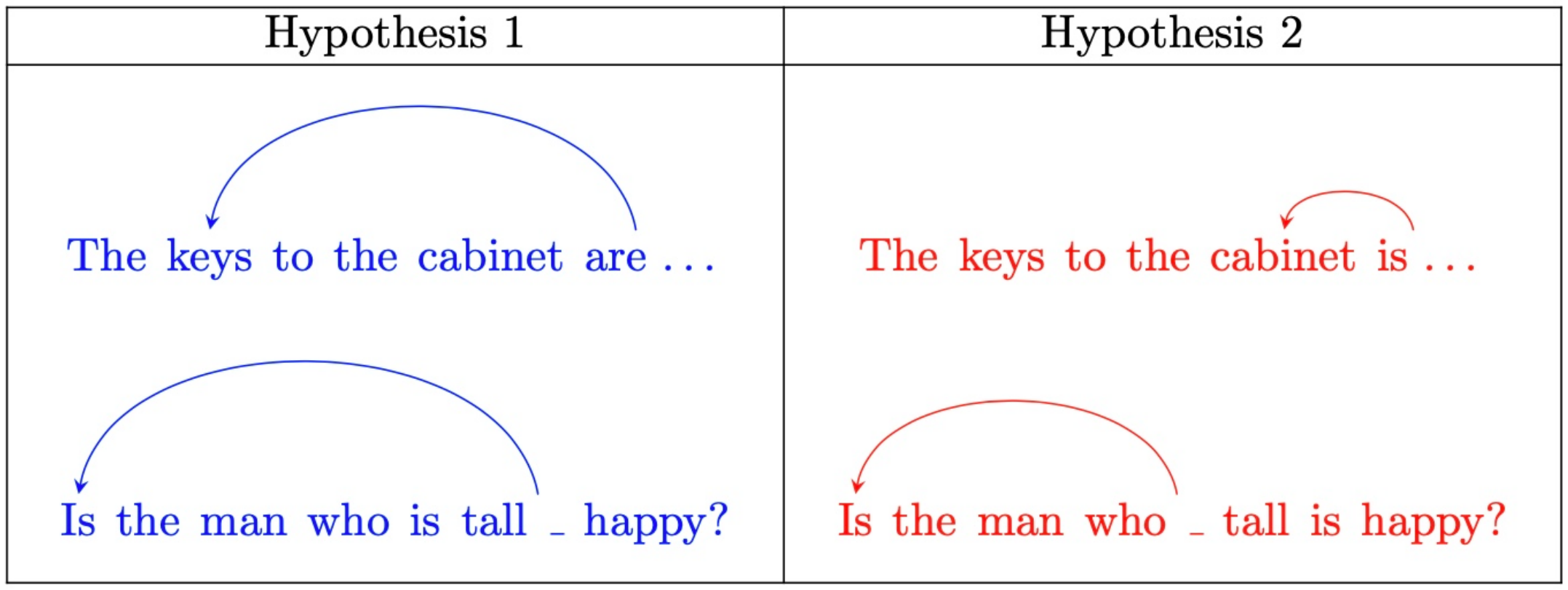

Figure 2: Illustration of the role of inductive bias in generalization. **A.** The black points represent data, and the lines represent generalizations that could be formed on the basis of the data. Both lines fit the data exactly, but make different predictions for new datapoints. Intuitively, the blue line seems like a better hypothesis. But the choice to prefer any single generalization over any other can only be a function of *inductive bias*—a preference for hypotheses which are in some sense 'simpler'— which is not a function of the data. **B.** Inductive bias in language. In black, possible corpus data from English. In blue, sentences which generalize from the corpus data on the basis of hierarchical structure. In red, sentences which generalize on the basis of linear word order. Intuitively, the blue sentences seem like natural generalizations, and the red ones seem unnatural, even though all are equally consistent with the data given. This phenomenon is claimed to be the result of innate biases specific to language (Chomsky, 1971).

Inductive bias shows up in several forms and under many names in the scientific literature. In linguistic theory, it appears perhaps first as the *evaluation procedure* in Chomsky (1965, pp. 30–48), a function which compares two different grammars that can generate an observed set of sentences (that is, two grammars that fit some learning data equally well), and ranks them in order of preference, in a way that captures how humans generalize beyond the data (Chomsky, 1965, p. 45). More generally, it appears as Universal Grammar (UG)—an "innate schematism of mind that is applied to the data of experience" (Chomsky, 1971, p. 28) that enables language learning and generalization; furthermore, UG is held to be domain-specific to language (not applying to any other aspect of cognition) and species-specific to humans (Huybregts, 2019). In this approach, (generative) grammatical theories such as Minimalism are hypotheses about the nature of UG (Chomsky, 1993; Adger, 2003). They are meant to precisely delimit what languages may exist and be acquired by humans.

This approach to linguistic theory serves the goal of *explanatory adequacy*: the idea is that a theory of grammar should not only capture which sentences are grammatical and ungrammatical in a particular language, but also that the theory encompass all and only the possible languages that we might actually find in the world (Chomsky, 1965, Ch. 1). Thus, if a grammatical theory seems to accommodate 'impossible' or unattested kinds of languages, then this would be considered evidence against that theory, because the theory is supposed to represent the universal and language-specific "schematism of mind" that a learner applies to language data.[7] This approach to explaining linguistic phenomena has been called "explanation by constrained description" (Haspelmath, 2009, pp. 384–385).

Naturally then, Universal Grammar has been the major focus of study in language learning, seen as the simultaneous solution to two different problems: (1) how children can learn language from inadequate data (the classic Argument from the Poverty of the Stimulus: see Pearl, 2022, for a review), and (2) why human language is the way it is: because Universal Grammar strongly restricts the set of possible languages.[8] The dream has been to come up with a formalism for linguistic description which captures human generalizations, is domain-specific to language, and arises from a genetic endowment

[7] For example, under this view, it would be justified to criticize a theory of grammar such as Head-Driven Phrase Structure Grammar (HPSG: Pollard and Sag, 1994; Sag et al., 2003) on the basis that it is Turing-complete, capable of generating any recursively enumerable formal language.

[8] Given the centrality of this reasoning to much of linguistic theory, one might expect it to be backed up with strong experimental evidence that humans cannot or do not learn languages which violate the putatively universal principles of human languages. This is not the case: there is only limited and ambiguous experimental evidence for hard formal limits on human language learning.

Among the most on-target existing studies is Smith et al. (1993), whose object of study was a man described as a polyglot savant living in a mental health facility. This individual and four control subjects (linguistics undergraduates) were tasked with learning artificial languages designed to be 'impossible' in three ways: (1) negation and tense are indicated by word order, (2) there is an agreement pattern judged to be impossible, and (3) the position of an emphatic marker is determined by a rule involving counting words. Results are not systematically reported, but seem to indicate that the polyglot was able to learn the 'impossible' word order and agreement rules (1) and (2), but not the rule for the emphatic marker (3).

Another commonly cited experimental work on this topic is Musso et al. (2003), who expose German speakers to Italian and Japanese sentences, either following the real rules of those languages, or following modified rules deemed to be linguistically impossible, for example placing a negation marker after the third morpheme from the beginning of a sentence. The result is equally accurate learning of the 'natural' and 'unnatural' languages. fMRI on the subjects shows that the real languages elicit activity in the left inferior frontal gyrus, while the unnatural ones elicit activity elsewhere. We believe the meaning of these results is unclear. Only a small number of languages and participants (all of whom were already native speakers of largely hierarchically-structured languages) were tested, the localization of syntax in the brain is still contentious, and the patterns of brain activity for the 'unnatural' languages might reflect a lack of practice with such patterns, rather than their impossibility or a qualitative difference between linear and hierarchical rules.

unique to humans (Chomsky, 1988; Hauser et al., 2002; Berwick et al., 2011)—thus in one fell swoop solving (1) and (2).

Under this logic, it follows that a good theory of language learning should be *restrictive*: that is, there should be languages that *cannot* be learned under the theory, and this restriction on the hypothesis space provides explanatory adequacy.[9] In terms of Figure 2, the intuitively bad hypotheses would be ruled out as simply unavailable as mental representations during learning (Everaert et al., 2015).[10] This approach to linguistic explanation has a pleasing elegance to it: learners *must* be restricted to learn properly, and we see that the variation in actual languages is restricted, therefore we can kill two birds with one stone by finding the right representational restrictions on learners.

### 3.2.2 The modern view on learning

The logic of inductive bias is sound. On a deep level, there really is no free lunch in learning, even deep learning (Mitchell, 1980; Wolpert et al., 1995; Baxter, 2000; Adam et al., 2019). Asking how much one can learn 'from the data alone' without inductive bias is like asking how close a runner can get to the finish line without saying where they started. It's just not a sensible question. However, developments in deep learning have forced revisions to conventional ways of thinking about how inductive bias arises.

The conventional wisdom was that a learning model must be restricted in terms of the hypotheses it can consider, so that it does not overfit (Bishop, 2006, §1.1). A relatively unrestricted model may memorize the training data, or find some other pathological solution which lurks in the depths of its vast hypothesis space, as in the red hypotheses in Figure 2. By restricting the hypothesis space, one provides strong inductive bias, so that pathological solutions can be ruled out and good generalization can be guaranteed. The logic is the same as Universal Grammar: explaining generalization requires that we restrict the set of hypotheses (grammars) that a model (learner) may entertain.

However, the empirical finding from deep learning is that *overparameterized* networks, which are more than flexible enough to memorize their training data, do generalize, often better than relatively restricted ones (Belkin et al., 2019; Zhang et al., 2021). In machine learning, overfitting has typically been measured in terms of performance on a held-out dataset, where one typically sees a U-shape curve as in the left part of Figure 3: as models become more flexible in terms of the hypotheses they can express, at first they generalize to the held-out data poorly (underfitting), then they generalize well (the sweet spot), then after some point the more powerful models learn poorer hypotheses (overfitting). And yet in the late 2010s it was discovered that if you keep adding model power, the U-shape curve starts to descend again, in a phenomenon called *double descent*, shown in Figure 3. This pattern suggests that learners continue to find good generalizations even *after* being flexible enough to memorize their training data.[10]

---

[9] See, for example, Kodner et al.'s (2022) criticism of Yang and Piantadosi's (2022) model of language learning as Bayesian program induction, a model which successfully learns grammars of various formal classes given small amounts of string input, thus addressing the Poverty of the Stimulus problem for formal structures. Although the model readily meets the challenge of inducing formal structure from strings, it has been dismissed by some in the linguistics literature because the same model could also learn grammars that are unlike human language.

[10] Although these are common examples, it is not clear exactly how generative formalisms rule out the unnatural hypotheses here. In particular, the languages implied by the unnatural hypotheses are context-free, just as much as the languages implied by the natural hypotheses. So these (string) languages could be generated from, for example, Minimalist Grammars (Chomsky, 1993; Stabler, 1997), since Minimalist Grammars generate a superset of context-free languages (Michaelis, 1998).

[10] In certain settings and with the right regularization, the best performance comes from models whose capacity is poised right at the point where training data and model parameters are balanced—the right regularization in these settings avoids the sharp spike in loss characteristic of double descent (Maloney et al., 2022, §4.2).

It seems that what appeared to be an insurmountable barrier turned out to be a kind of phase transition in how learning works. What looks like overfitting is, counterintuitively, ameliorated by adding even more parameters. Thus, one sees a new kind of generalization emerging, where less restricted learners have superior performance in learning the right generalizations.

These findings came as a surprise to statistical learning theorists (Belkin et al., 2019; Zhang et al., 2021) and have triggered an ongoing field-wide effort to rethink learning theory or to show how these unexpected findings are compatible with existing theory (for example, Poggio et al., 2020a; Martin et al., 2021; Martin and Mahoney, 2021; Kuzborskij et al., 2021; Henighan et al., 2023; Attias et al., 2024). These results should also make us rethink language learning and the role of restrictive formalisms in linguistic explanation. It is simply not the case that proper generalization can only come from learners who are sharply restricted to small hypothesis spaces, nor even that there is a correlation between restrictedness and proper generalization. All the successes of modern machine learning are in contradiction to the view that generalization must come from hard restrictions on the hypothesis space.

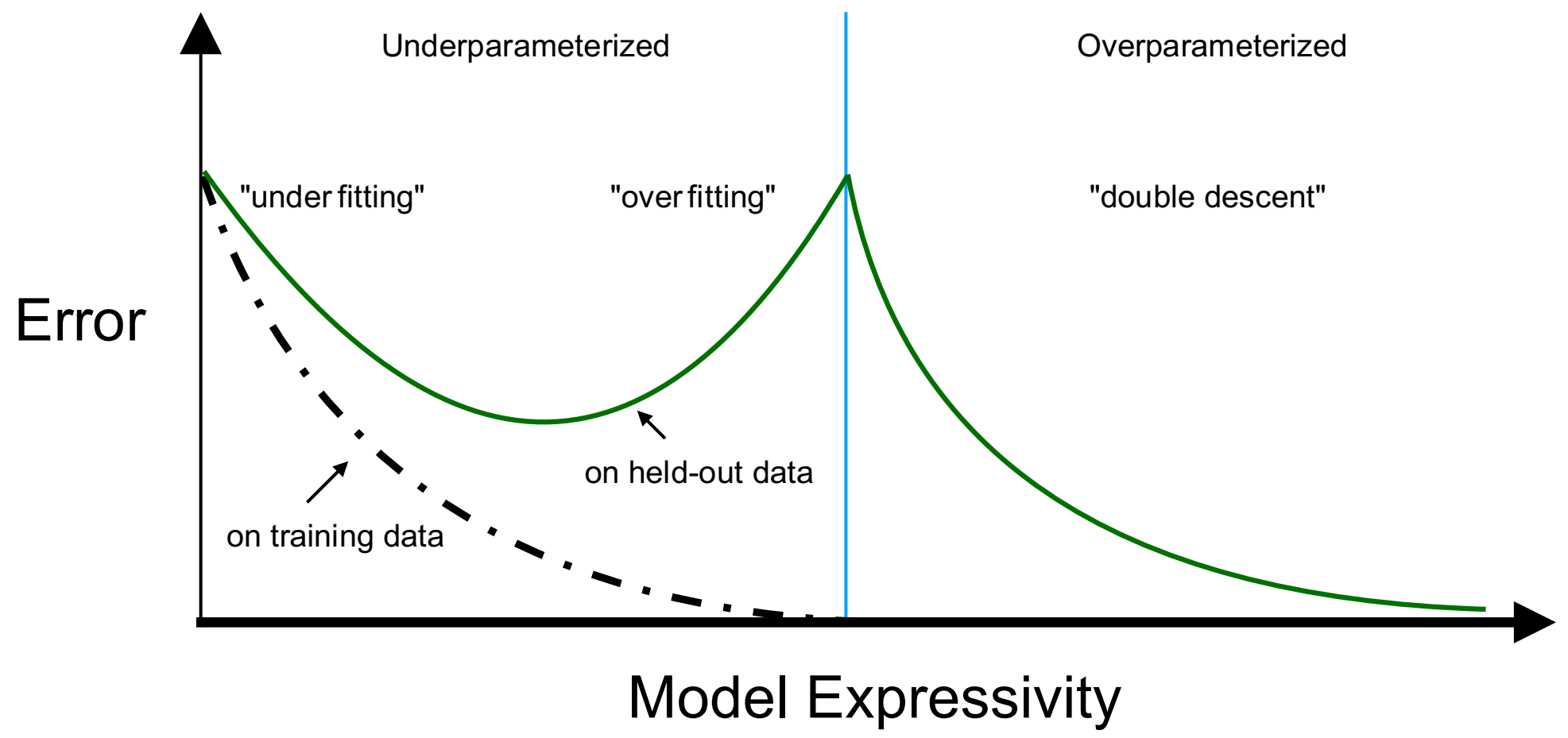

Figure 3: Expected error of a learning model on training set and test set as a function of model expressivity (the variety of hypotheses that the model can express). To the left of the blue line, we have the classical picture from statistical learning theory. In the *underfitting* regime, increasing model capacity reduces training and test error, up to a point where one enters the *overfitting* regime, in which increasing model capacity causes a decrease in error on the training data but an increase in error on held-out data. This overfitting phenomenon is intuitively due to the model gaining the capacity to memorize the training set without forming generalizations that would be useful on the test set. The goal of model fitting in this view is to find the *sweet spot* that minimizes test error, often accomplished through methods such as deliberately reducing model capacity or early stopping. However, modern deep learning has revealed that as model capacity increases *beyond* the point where the model can memorize the training data (the *interpolation threshold*, in blue), we enter a new *overparameterized* regime, where test error decreases, due to soft simplicity biases in the learner. Figure based on Belkin et al. (2019).

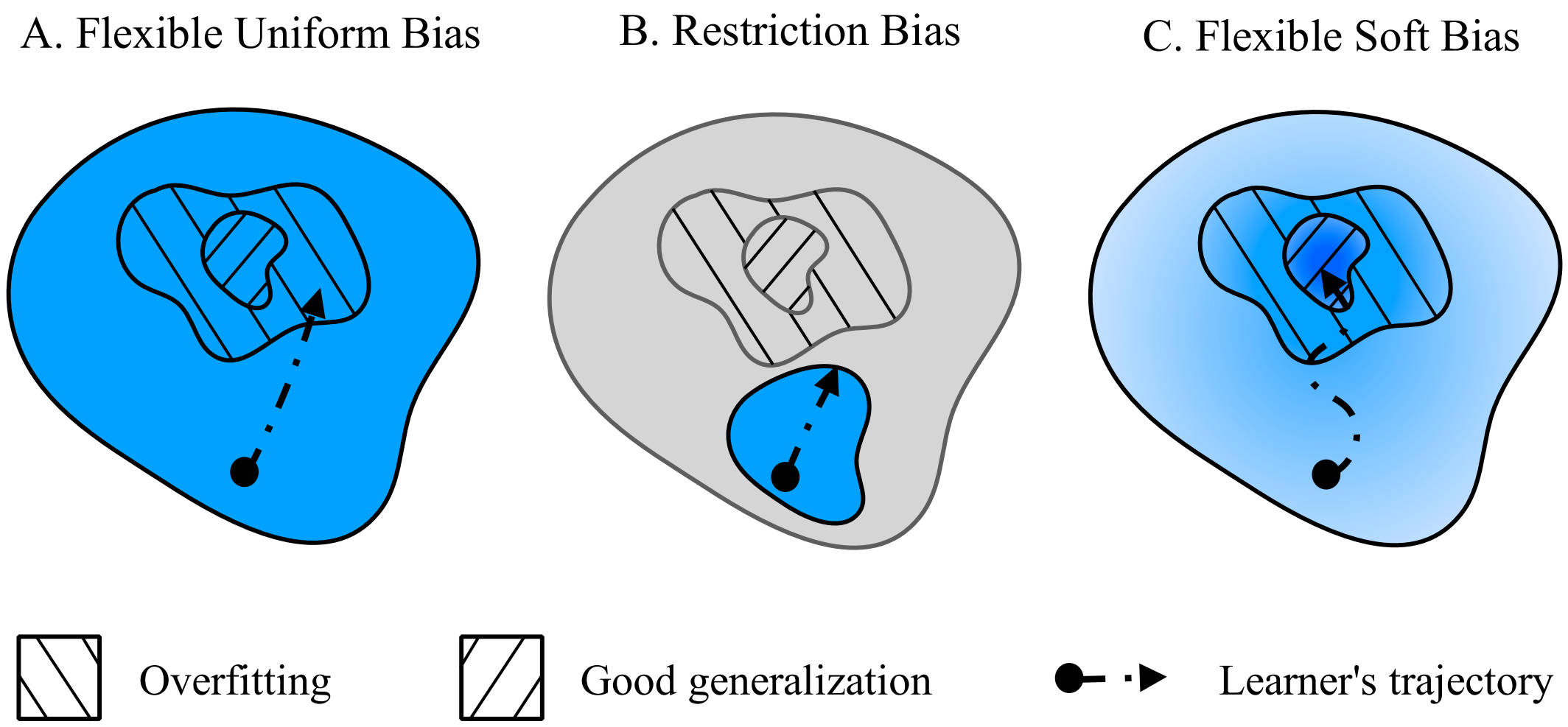

Figure 4: Illustration of the logic of a soft inductive bias within a flexible hypothesis space, adapted from Wilson (2025). Figures show a hypothesis space for a learner, including regions of bad generalization (overfitting) and good generalization. The learner starts at the dot and, throughout learning, moves in the direction indicated by the arrow. **A.** Given a uniform bias over a highly flexible, relatively unrestricted hypothesis space, the learner is likely to overfit and form pathological generalizations. **B.** One strategy to impart useful inductive bias to a learner is to restrict the hypothesis space. However, this may miss the good generalizations. **C.** A better strategy, which is part of what underlies neural network success, is to keep a highly flexible, relatively unrestricted hypothesis space, but impart a soft simplicity bias which guides the learner towards the good generalizations.

But how can this be, if the idea of inductive bias is right? How did the conventional logic go wrong in practice? The key mistake was the conflation of model power with inductive bias (Hubinger, 2019). The modern empirical finding is that *more* flexible learners have *stronger* biases toward "simple" hypotheses (Huh et al., 2024): that is, although they do not restrict the range of hypotheses that *can* be considered, they *do* impose a soft notion of simplicity on those hypotheses. The logic is illustrated in Figure 4, following Wilson (2025). Furthermore, the source of this simplicity bias in neural networks and related systems is not necessarily related to the hard limits of their expressivity. The simplicity bias in neural networks seems to be toward functions that are nearly linear or simple in other ways (Valle-Perez et al., 2018; Hahn et al., 2021b), likely as a result of the dynamics of gradient descent on the loss landscape induced by the model (Poggio et al., 2020b; Pezeshki et al., 2020; Merrill et al., 2021; Hahn and Rofin, 2024).

In machine learning, these discoveries have catalyzed a change in focus, away from models whose architecture and representations are tailor-fit to their domain, and toward models that learn quickly in relatively unrestricted hypothesis spaces (Sutton, 2019). The zeitgeist is now strongly against using domain knowledge to restrict the behavior of learners, even when researchers have a strong sense of what the relevant domain knowledge is.

#### 3.2.3 The upshot for linguistic theory

These surprising results from deep learning mean that the research program of achieving explanatory adequacy through restricted learners is deprived of its apparent logical inevitability. Previously, the logic was: learners *must* be formally restricted, and we see that human languages *do* have universal patterns, so maybe these necessary restrictions create the universal patterns. Nowadays, if this approach to linguistic explanation is to be maintained, the logic must be: learners have hard formal restrictions *even though this is not necessary and may be harmful for learning*, and these restrictions create language universals. It is still a viable hypothesis, but it loses its elegance.

Furthermore, even if language-specific innate inductive biases in humans really are the key to the structure of human language, these inductive biases might not be expressible in terms of a categorical symbolic formalism or a sharply limited hypothesis space for learners. Inductive biases in modern neural models are soft and seem to arise from a complex interplay of training dynamics, objective function, and model architecture, with the hard limits of model expressivity playing a relatively minor role.

Overall, "explanation by constrained description" no longer seems so explanatory, at least not without further stipulations. Language learning and language universals may well be better captured by a highly flexible, less constrained formalism for linguistic representation—one which on its own could capture non-linguistic patterns as well as natural linguistic ones—paired with a soft, quantitative simplicity metric that captures learning dynamics or functional pressures on language. Learning models with explicit simplicity biases exemplify this approach (for example, Hsu et al., 2011; Perfors et al., 2013; Rasin et al., 2021; Lan et al., 2022).

This is not to say that inductive biases are no longer important in linguistics and language learning, nor that investigating inductive biases of humans and neural networks is not important: we will have much more to say on this in Section 4.4. Far from demoting inductive bias as a concern, language models open up the range of (possibly innate) inductive biases that we might look for in humans. The main point for linguistic theory is not to demote the importance of the learning problem, but to reshape it: away from explanation by constrained description, and toward a broader landscape of approaches and hypotheses.

#### 3.2.4 The question of data quantity

It is still the case that, whatever language models learn about language, they do it using orders of magnitude more linguistic input data than human children are exposed to (Yedetore et al., 2023; Warstadt et al., 2023). Moreover, the learning trajectories of models and humans continue to show systematic differences (Chang and Bergen, 2022; Evanson et al., 2023; Constantinescu et al., 2025). Taken together, these results suggest meaningful differences between learning in models and humans, both in terms of data requirements and patterns of learning. A form of the Poverty of the Stimulus argument is still alive in the form of the observation that even if neural networks acquire linguistic structure, they do not do so on the basis of the same amount and kind of data that a human child learner gets (Lan et al., 2024).

As reviewed above, it has been suggested that formal linguistic structure provides the key inductive bias that makes learning from small amounts of data possible. However, there is no evidence that linguistic structure is the major missing ingredient to close the gap between models and human children.[11] There have been many attempts to inject linguistic structure into neural models in various ways; some of the earliest successful deep learning approaches to language understanding were based on architectures that performed recursive computation in hierarchical parse trees (Socher et al., 2011, 2013). These approaches do succeed in promoting the kinds of structural generalizations that humans find intuitive, with the result that language learning from data is somewhat more sample efficient (Dyer et al., 2016; Futrell et al., 2019b; Wilcox et al., 2019b; Kim et al., 2019; Papadimitriou and Jurafsky,

[11] Indeed, the analysis by Mollica and Piantadosi (2019) suggests that syntactic structure makes up only a very small portion of the information necessary to learn a language.

2020, 2023; Nandi et al., 2025). However, in terms of absolute performance, these approaches are still far from closing the gap between humans and neural networks.

It is entirely possible that the key to faster and more human-like learning is not particular built-in formal constraints, but rather *more flexible* architectures (such as Kolmogorov–Arnold Networks: Liu et al., 2024), or biases towards domain-general compositional reasoning (McCoy and Griffiths, 2023; Yang and Piantadosi, 2022), or different training regimes (Murty et al., 2023), or the incorporation of multimodal data which provides rich side information about the structure of the environment that is being described in language (Wang et al., 2023), or domain-general bounded-rational approaches to generalization such as the Tolerance Principle (Belth et al., 2021; Payne et al., 2021; Kodner, 2022).

Overall, we believe the shoe is on the other foot for those who believe that an innate bias toward a particular formal linguistic structure is the factor that enables human learning from small amounts of data. Formal linguistics has not presented an alternative model with the demonstrated practical language-learning capabilities of neural models. The direction of developments in machine learning suggests that the gap between human and machine learning is more likely to be closed not through built-in domain-specific formal restrictions, but rather through more powerful domain-general learning algorithms.

The difference between human and neural network learning also raises the possibility that humans and neural networks are just so different that one is not informative about the other (Kodner et al., 2023). We believe not. First, as argued above, the success of neural networks weakens *logical* arguments that language cannot be learned without domain-specific formal constraints on language, and in general changes how we think about the role of formal restrictions in learning. Second, *even if* standard neural network training methods are not able to acquire linguistic structure on the basis of developmentally realistic data, the representations that they do eventually acquire based on more data are still informative about how language might be represented and processed in the brain (more in Section 4), even if the networks do not arrive at these representations along the same trajectory as the human child.

## 3.3 Language models and linguistic traditions

While language models may seem to involve concepts that are foreign to linguistics, they emerged in part from certain intellectual traditions of the study of language—although these traditions are not necessarily the ones that have been dominant in linguistics departments. Below we discuss the relationship between schools of linguistics and the development of language models and related language technologies.

### 3.3.1 The generative tradition of linguistics

It is rare in the history of science for a scientific theory to turn out as disconnected from a corresponding engineering application as formal generative linguistics has turned out to be for language models. We believe this has happened primarily because of a difference in goals between generative linguistics and language modeling, with the generative tradition taking a narrow view of its goals.

A historical parallel is informative. In the early 19th century, while Newtonian mechanics provided a strong basis for parts of mechanical engineering, it did not on its own seem to provide a theory that answered pressing questions about the increasingly complex machines emerging from the Industrial Revolution, in particular steam engines. Carnot (1824) developed an effective theory of such engines using a new ad-hoc concept of 'moment of activity', which eventually developed into the idea of entropy (Clausius, 1865). This concept had to be discovered through engineering because the focus of purely theoretical physics was on understanding fundamental mechanics through mathematical methods of increasing elegance. Explaining the behavior of complex machines—which behavior certainly must reduce to more basic mechanics anyway—was not on the agenda. And yet from the practical project of understanding steam engines emerged a family of concepts that some theoretical physicists now view as more fundamental than even matter itself (Wheeler, 1989).

Turning back to linguistics: the reason that generative linguistic theory did not turn out to be helpful along the path toward language models was due to a similar narrowness in focus, which led it away from considering complex systems for dealing with language.

Language models are, at bottom, models of the stream of language that is produced and comprehended by humans. The relevant theory is the theory of language use, production, comprehension, learning, and of human cognition more generally. Generative linguistics, on the other hand, has not focused on how to build a machine that produces and comprehends language, but rather *how to build a language*, conceived of as an abstract mental structure that gives rise to a mapping between meaning (that is, a logical form or conceptual–intensional representation) and form (that is, a phonological form or sensorimotor representation) (Chomsky, 1995, 2005; Adger, 2003; Hornstein et al., 2005).[12] Furthermore, it was claimed that this kind of analysis must take center stage in the science of language, preceding any analysis of more complex systems for language processing, use, or learning, since these systems must operate in ways that make reference to the abstract structures of language (Chomsky, 1965, Ch. 1).

Given this focus, generative linguistics *has* had a huge influence on the engineering field that works on how to build languages: programming language design. Indeed, the basics of all modern programming languages are based on formal linguistic theory, incorporating explicit (often contextfree) grammars and parsers, as a way of linking a stream of text to a representation of a computation. Programming languages are, perhaps, exactly the 'esoteric dialects' spoken by 'elaborate machines' in the Nabokov quote that starts this section.

While this narrowness of focus was useful on its own terms, it has limits. There are potentially fundamental insights to be gained now from the analysis of messy, complex, practical systems, just as happened in physics, and as is now happening in linguistics.

### 3.3.2 The statistical tradition of linguistics

In fact, there is a long tradition of statistical approaches to language, and these traditions were deeply involved in the early development of language models.

Perhaps the most well-known such contribution is *distributional semantics*: the idea that the semantics of a word is related to the distribution over contexts in which that word appears, often cited to Firth (1957, p. 11), and developed more systematically by Harris (1954). This idea originates from the school of structuralist linguistics (Saussure, 1916; Bloomfield, 1926). Structuralist linguistics had as one theoretical aim the development of *discovery procedures*, which were formal procedures that could be applied to bodies of text in order to discover (and even *define*) linguistic structures (Harris, 1951). The most well-developed of these discovery procedures were statistical in nature. For example, Harris (1955) developed a theory of words and morphemes based on statistical co-occurrence patterns, including a procedure for discovering morpheme boundaries by effectively calculating transitional probabilities, an idea taken up again much later in the psycholinguistics of language learning (Saffran et al., 1996), and closely related to tokenization methods such as byte-pair encoding (Shibata et al., 1999; Tanaka-Ishii, 2021, Ch. 11). At issue was the relationship between grammatical structure and the observable statistical structure of a corpus of language—one of the core questions that linguists working on language models are interested in again today.

[12] Early generative linguistics went further in its goals, claiming that the structures identified through generative analysis would not only provide a characterization of language, but also be a necessary and explicit part of any theory of how language could be learned and processed (Chomsky, 1965, 1971; Bresnan, 1982). See Stabler (1983) for discussion of what he sees as confusion within the field as to whether or not generative linguistic theories are intended to be theories of the representations used by the brain during processing, as opposed to theories that constrain possible language. He concludes that Chomsky and others often conflated theories of grammar and theories of mental representation and processing, but that this position was not well grounded and that "it seems unlikely that linguists and psychologists really want to claim any such thing".

Harris's work is an intellectual precursor to modern language models. However, in practice, the statistical structuralist tradition represented by this work was largely supplanted in American linguistics departments by the generative school, which sprang from Chomsky's (1957) arguments that distributional statistics were irrelevant to linguistic structure and that discovery procedures were a distraction from the putatively 'core' questions of linguistics.

Nevertheless, the statistical analysis of language continued under the heading of usage-based approaches (Bybee and Hopper, 2001), in which the key to language is not a set of underlying formal rules but emergent properties based on the statistics and dynamics of language use (from which rules or rule-like behavior might emerge). Traditions in typological syntax (Greenberg, 1963; Dryer, 1992), functionalist syntax (Comrie, 1989; Keenan and Comrie, 1977), construction grammar (Croft, 2001; Goldberg, 2009), probabilistic modeling (Bresnan et al., 2001, 2007; Christiansen and Chater, 2016), and evolutionary linguistics (Kirby and Hurford, 2002) have all carried the banner of a thoroughgoingly statistical approach to language. Earlier arguments about the relationship between statistics and structure, and the value of probabilistic methods, mirror many of the points we are making here (Abney, 1996; Pereira, 2000; Manning, 2003; Bresnan et al., 2007; Lappin and Shieber, 2007; Norvig, 2012).

These statistical traditions of linguistics played a catalyzing role in the birth of neural networks and LMs. The connectionist framework from which neural LMs emerged overlapped with the statistical tradition in linguistics, often in conflict with the anti-statistical tradition. A major locus of this early work using neural networks was about the English past tense, with Rumelhart and McClelland (1987) developing a neural model for forming the past tense from present tense words, triggering much debate (Pinker and Prince, 1988). Elman, a Linguistics PhD, developed an early recurrent neural network (RNN) sequence architecture (1990a) in order to solve the problem of finding linguistic structure in time, a problem motivated by findings in linguistics and psycholinguistics (Frazier and Fodor, 1978). This basic architecture still underlies many language models (Feng et al., 2024a). Chris Manning, a pioneer in neural and probabilistic models, was trained as a linguist (Manning, 1995), and argued for a statistical approach to syntax (Manning, 2003), before going on to work on influential neural methods for performing linguistic tasks (e.g., Pennington et al., 2014).

The history of the field and the dominance of the generative tradition made it such that these figures are sometimes seen as 'not linguists'. But the traditions they emerged from were fundamentally concerned with questions about human language and cognition and belong under the heading of linguistics. To be clear, we do not think that the success of modern language models *unequivocally* supports or refutes any particular intellectual tradition in linguistics. But we do think that the narrow and exclusive theoretical focus of generative linguistics, coupled with its relative dominance within American linguistics departments throughout the late 20th century, caused tragic missed connections in intellectual history, and has left the field of linguistics diminished compared to where it could be.

Progress in language science will not come from hyperfocus on any one goal, nor from any one theory of what language is, nor from one framework for understanding linguistic phenomena. Rather, the science of language needs to draw on a plurality of perspectives, from a wide variety of disciplines. In the current moment, that means leveraging the explosion of intellectual creativity springing forth around language models. Just as the concept of entropy arose from ad-hoc analysis of complex machines and ended up revolutionizing fundamental physics, it is possible and even likely that ideas based on language models will revolutionize linguistics.

# 4 Where does that leave the science of language?

> "Trying to understand perception by studying only neurons is like trying to understand bird flight by studying only feathers: It just cannot be done. In order to understand bird flight, we have to understand aerodynamics; only then do the structure of feathers and the different shapes of birds' wings make sense."
>
> Marr (1982, p. 27)

If neural models make us rethink some aspects of linguistic theory, where do we go from here? What is the status of the formal structures discovered through linguistic analysis, if they don't have to be innately latent in the human brain? We believe that linguistic structure is as real as it ever was, and that the LM revolution presents us with important new ideas and methods for studying language. Taking a statistical and information-theoretic approach to language brings these possibilities into focus. We also consider ways in which LMs might give us reason to update our linguistic theories to be more gradient, usage-based, and functionalist, by serving as a proof-of-concept system that implements ideas that were previously hard to formalize.

## 4.1 Linguistic structure is real

LMs are proof-of-concept that systems can process language without having linguistic structure hardwired in. But that doesn't mean it isn't real. We hold that linguistic structure is a *real pattern* in the sense of Dennett (1989): it provides a compressed and useful representation of important aspects of language. To explain or describe linguistic phenomena without reference to linguistic structure, perhaps in terms of some more reductive neural theory, would be hopelessly complex — and would make understanding LM behavior more difficult (see Nefdt, 2023, for a worked-out theory of linguistic structure as real patterns).

Figure 5 shows a schematic of how this account works for a phenomenon like subject–verb agreement. Modern language models are proficient at this task, proficient enough that their behavior looks rule-like. But, given the dynamic and chaotic nature of neural networks, it is likely that there are messy and heterogeneous structures and processes that give rise to this rule-like behavior in models. These structures might only very noisily map onto linguistic categories like "grammatical subject" or "number agreement". One reaction to that might be to conclude that linguistic theory was wrong or misguided—that it seemed like there was such a thing as "grammatical subject" that was crucial for verb agreement, but that actually the underlying reality was more complicated and now we can do away with all these rules and replace them with the messy, complicated neural system instead.

This hyper-reductive conclusion would not be productive. A theory of language which points to complex neural circuits for explaining the data of subject–verb agreement, without referring to something about the concept of grammatical subject, would be like an example from Dennett (1989, Ch. 2): imagine aliens who can perfectly predict human behavior using atomic physics (that clump of human atoms will move towards that coffee shop) but without understanding human beliefs and desires (that the human wants coffee and believes that the coffee shop will provide it). Lacking concepts of humans' attitudes, the aliens' theory of human behavior is perhaps perfectly accurate and reductive, but comes "at enormous cost of cumbersomeness, lack of generality, and unwanted detail" (Dennett, 1990, p. 189). Dennett concludes that the concept of belief is a *real pattern* in that it provides an abstraction that supports prediction and counterfactual reasoning based on coarse-grained data. The idea is that any abstraction that enables simple prediction, compression, and causal modeling in this way is a real pattern, fully deserving the epithet *real* even if there is a more reductive theory that lacks the abstraction. Similarly, even if it turned out we could explain linguistic behavior without recourse to anything recognizable as linguistic structure, it would come at the same cost of “cumbersomeness, lack of generality, and unwanted detail”.

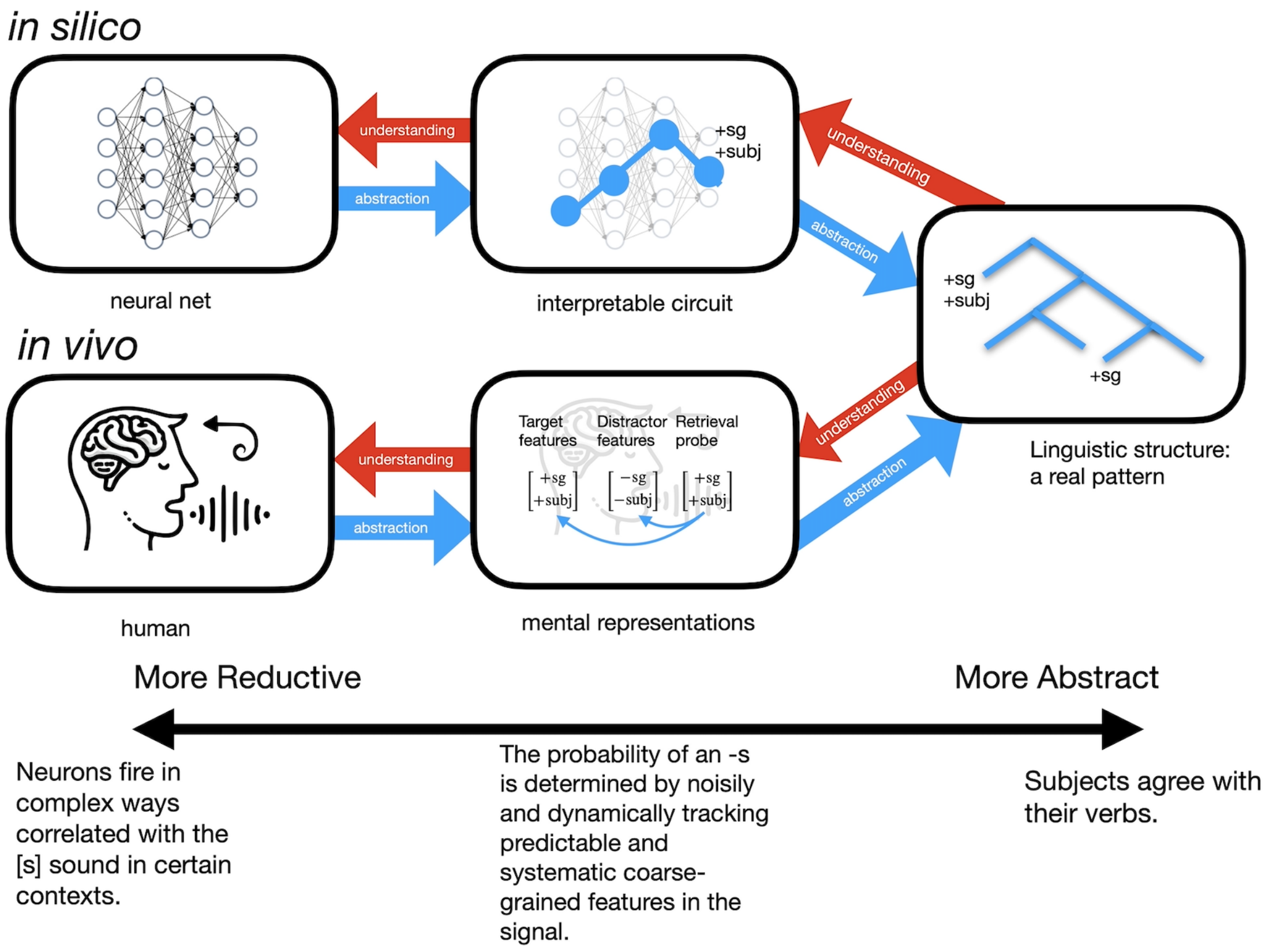

Figure 5: Schematic for linguistic structure as a *real pattern*, a (potentially leaky) abstraction that supports prediction and counterfactual reasoning based on coarse-grained data. In humans (*in vivo*), the complex physical, biological, and social processes associated with language learning, use, and knowledge may be abstracted into psycholinguistic theories describing operations on mental representations. In turn, the behaviors of these mental representations may be interpretable in terms of coherent linguistic structure. In neural networks (*in silico*), the complex patterns of weights and activations may be abstracted into interpretable circuits that approximately compute linguistically meaningful features like whether a word is singular and a grammatical subject. In turn, those interpretable circuits may be interpretable in terms of a larger abstraction of a coherent linguistic structure. Thus linguistic structure is a real pattern at the highest level of abstraction, which allows us to understand language as it is implemented by a human or by a neural network.

We join Dennett in his conviction that real patterns are, well, real. As such, we are not eliminativists about grammatical structure, nor do we believe that this kind of eliminativism would be scientifically productive. Some in the generative tradition adopt a similar view, contra the idea that every posited linguistic structure must be in the head. Adger (2022) writes: "Generative grammarians think of the posits of their theories in much the same way as physicists think of quarks or particles in wave form: They are the best explanations of the phenomena, though we do not know exactly what they correspond to in the world we can currently observe." LMs are a promising avenue for figuring out what these explanations do, in fact, correspond to in the world.

Indeed, the idea of real patterns can help explain the success of language models. Language models are trained to predict tokens in context. From an information-theoretic perspective, this means they are finding simple representations for linguistic input (Chater and Vitányi, 2007), and in particular, compressed representations of past input that let them predict future input (Tishby et al., 1999; Shalizi and Crutchfield, 2001; Bialek et al., 2006; Still, 2014; Tishby and Zaslavsky, 2015). Insofar as linguistic structure is real, it is part of the representation that enables this compression.

A similar point is made by Wolfram (2023), who connects the success of LMs with the reducibility of computational processes. Some computational processes are chaotic: there is no good way to systematically predict the digits of pi other than actually running the computation and learning what the next digit is. But other processes that *appear* complex have learnable patterns that can be used to give simpler, compressed representations. For instance, $\frac{3227}{555}$ is 5.8144144144... where the 144 repeats forever. This repeating structure means that we could give a compressed representation of what the millionth digit is without having to do additional computation. Wolfram suggests that the success of LMs proves that human language has structure that enables this kind of compression. This compressibility is responsible not only for LMs' success, but for the fact that it is possible to construct linguistic theories. Compressibility *is* structure.

## 4.2 What LM interpretability can tell us about human language

We do not mean to suggest that linguists and cognitive scientists should identify compressible patterns in behavior, call it a real pattern, and declare victory. Perhaps *the* grand challenge of all of cognitive science is to uncover the internal mechanisms in neural networks and brains that underlie and implement behavioral patterns (Smolensky, 1988). Finding these internal mechanisms is not a prerequisite for the reality of linguistic structure—the mechanisms may be irreducibly complex. But when mechanisms *are* found in neural networks, they can be informative not only about linguistic structure itself, but also about how language processing and learning may be implemented in the human brain.

**Interpretability in an idealized thought experiment** Consider a thought experiment, using the linguistic concept of c-command (a particular relationship between elements in a linguistic tree which has been posited to be important for a variety of structure-sensitive linguistic patterns). In this thought experiment, we get embedded representations for each word and use an interpretability method to get a representation of the pairwise relationship between each pair of words in each sentence. We analyze these word pair representations, and we find that, in layer $n$, neuron $m$ patterns in an interesting way. If and only if the value of that particular neuron is positive, then the first word in the pair c-commands the second word. If it's negative, then we know that it doesn't. Further experimentation reveals that this pattern is remarkably consistent. (N.B.: We want to be clear that no experiment would ever give such clean results; LMs just don't represent anything in such human-interpretable ways.)

If we did discover a result like this, we would be justified in drawing several conclusions. First, the ease with which we extracted the c-command relationship would be compelling evidence that, for performing linguistic tasks in English, understanding which words c-command which other words is useful — so useful that the model learned it emergently. Second, we would conclude that learning the c-command relationship is possible with relatively little built-in language-specific bias. We would thus update our credence that the learning of complex syntactic relationships like c-command is possible without a built-in Universal Grammar. Third, we could use this method to test various competing theories about which kinds of structural relationships should be more or less prominently represented in models. Fourth and most importantly, we could use these results to try to work out how an abstract structural relationship like c-command can be represented in the weights of any system which, like the brain, is composed of cascades of activating neurons. Such explorations would be informative as proof-of-concept *in silico* models of how the brain might process language.

If we *didn't* find the c-command relationship in our model representations, that's a trickier scenario. We might simply not have used the right methods to find c-command, even though it's robustly represented. Or it might turn out that, while c-command is critical to human processing, LMs process language sans c-command. Or it might turn out that neither humans nor models need to represent something like c-command to demonstrate proficient linguistic behavior. Even in this negative case, though, the results could be fruitful for generating hypotheses and stimulating further inquiry, grounded in data and requiring precise formulations.

Thus, while the linguistic conclusions we can draw from LM interpretability experiments require care, it should be relatively uncontroversial that the idealized c-command experiment above would be linguistically interesting—even while recognizing a wide variety of opinions as to whether the current slate of results from LM experiments *are* linguistically interesting.

**Interpretability in the real world** Our c-command example is not realistic: like all complex systems, language models don't learn and represent information in neat and interpretable ways. Rather, it takes a lot of work to ask questions about how LMs represent parts of speech (Tenney et al., 2019) or grammatical dependencies (Hewitt and Manning, 2019). And the answers that come out are messy.

But we are optimistic. The extent to which LMs are "blackboxes" is now overstated because of major advances in the field of neural network interpretability. For example, as discussed in Section 2.2, researchers now have a grip on how neural systems use the geometry of word embeddings to represent syntactic relations among words (Hewitt and Manning, 2019; Chi et al., 2020; Eisape et al., 2022; Diego-Simón et al., 2024).

More recent techniques have shown, in a mechanistic way, exactly how artificial neural circuits can be used to perform higher-level computations (Lindsey et al., 2025), finding neural circuits and induction heads in Transformers (Elhage et al., 2021) or using sparse autoencoders to find key features in model representations (Huben et al., 2023). These techniques often depend on *causal* manipulation—showing that, if particular parts of the neural network are perturbed or patched in particular ways, the output is affected in predictable ways (Geiger et al., 2023). For instance, Wang et al. (2023) shows how it is possible to trace a circuit that controls the completion of an object (e.g., "When Mary and John went to the store, John gave a drink to ." where the intended answer is "Mary"). Many papers have now used these techniques to *causally* study what parts of a network are responsible for particular kinds of complex linguistic behavior like grammatical number agreement (Lasri et al., 2022; Finlayson et al., 2021; Mueller et al., 2022; Lakretz et al., 2019), verb conjugation (Hao and Linzen, 2023), animacy processing (Hanna et al., 2023), and various kinds of long-distance dependencies (Arora et al., 2024). These techniques demonstrate how complex linguistic behavior can actually be implemented mechanistically in a neural system. These mechanisms are strong hypotheses for how syntactic relationships are represented in the human brain as well.

It is our hope—and also our prediction—that interpretability techniques will make it possible to close the gap between linguistic theory and implementation (mechanistic interpretability is a big part of "the way forward" per Millière and Buckner, 2024). This broad line of work, exploring symbolic representations in systems that solve genuinely interesting linguistic tasks, has started to make good on the promise of Smolensky's (1988) prescient vision "in which traditional and connectionist theoretical constructs collaborate intimately to provide an understanding of cognition".

## 4.3 LMs are a proof of concept for gradient, usage-based theories of language

The real patterns account leaves open the possibility that a wide variety of linguistic theories can be fruitfully combined with LMs. But there is a separate question raised here, which is whether the LM era should push us towards any particular theory for explaining language. We believe that the success of LMs strengthens the case for existing usage-based language theories based on gradient representations. By their very nature, such theories have been difficult to formalize to the extent that one could, for example, generate or process sentences with them. LMs serve as model systems for how these ideas can be implemented, and as proof of concept that they can work.

**Word Meanings** Traditional views in semantics often posit that words have discrete meanings and can be combined using systematic rules (Heim and Kratzer, 1998). In contrast, a distributional view of meaning assumes meanings are best captured by usage patterns (Potts, 2019; Goldberg, 2019; Erk, 2012; Baroni and Lenci, 2010; Baroni et al., 2014; Elman, 1990a), an intuition now operationalized inside language models as continuous high-dimensional vectors that encode the contexts in which a word appears, resulting in a rich web of lexical relationships represented within the geometry of the word vector space (Mikolov et al., 2013). If we think of the meaning of the word *fire* as being represented by its embedding in context, we find that the meaning of the word is slightly different in every context in which it occurs, but clustering together in space (campfires in a distinct cluster from a building fire even though these meanings are in some theories often lumped in as the same "meaning"; Chronis and Erk, 2020). As such, LMs show that we might not need a strict separation between core meaning and a richly context-dependent and pragmatics-laden meaning in use. LMs give us a way of formalizing the intuition that meaning might be gradient, fuzzy, and usage-based in a way that works in practice, not just in theory.

**Syntactic Categories and Grammatical Roles** LMs seem to be able to perform tasks that require knowledge of syntactic category and grammatical role, such as generalizing part of speech when encountering a novel token (Kim and Smolensky, 2021). One might then be tempted to ask: what are the syntactic categories an LM uses? And we can, for instance, probe for grammatical category in models and see whether the categories that emerge correspond to ones posited by linguists (Chi et al., 2020; Papadimitriou et al., 2021)? A key finding of this work is that, while there are natural clusters in part of speech and grammatical role, such clusters are necessarily fuzzy and gradient. For instance, Papadimitriou et al. (2021) found robust representations of subjecthood. But they were graded, with intransitive subjects and passive subjects less "subject-y" than the subjects of transitive active verbs. These representations could be discretized to make judgments like "BERT represents this word as a subject but not this one", but this discretization would be lossy. This graded notion of syntactic category and grammatical role has a long history in functional linguistics (Ross, 1972) and, as such, can be useful in giving us a new way to think about claims that word classes are not stable across languages (Haspelmath, 2012) and that grammatical categories exist in graded hierarchies (Comrie, 1989; Keenan and Comrie, 1977).

**Compositionality** While traditional theories have emphasized the compositional nature of language, the construction-based approach has emphasized that not all (or even most) language can be explained strictly compositionally (Goldberg, 2009). This position argues for a graded notion of compositionality: knowing what "green tea" is requires more than just knowing the meaning of "green" and the meaning of "tea"—even though this isn't strictly an idiom in that the meanings of "green" and "tea" are still relevant. There has been a lot of work on whether LMs can handle strict compositionality (Lake and Baroni, 2018; Kim and Linzen, 2020; Russin et al., 2024), as well as work on how LMs handle more idiosyncratic constructions (Mahowald, 2023; Misra and Mahowald, 2024; Weissweiler et al., 2023;

Tseng et al., 2022; Devlin et al., 2019). Neural models can combine the best of both worlds in that they can maintain multiple levels of representation at once (Baroni et al., 2014). They don't have to decide whether "green tea" is fully composed or fully stored: it can be a little of both. They can even be used to measure the degree of compositionality quantitatively (Rathi et al., 2021; Socolof et al., 2022).

**Separation of Linguistic Layers** Traditional linguistic theory has often posited a cognitive separation in processing between linguistic layers (phonology, syntax, semantics, pragmatics). But evidence from psycholinguistics (e.g., Tanenhaus et al., 1995; Shain et al., 2024; Fedorenko et al., 2020) suggests these layers are not processed in neat discrete stages but are richly integrated. LMs do seem to have some separation such that, for example, lower layers of the model handle morphology with higher layers handling syntax and semantics (Tenney et al., 2019). However, the nature of their architecture is to share information in hard-to-untangle nonlinear ways. So their functional separation is not strict but fuzzy and gradient—similar to psycholinguistic evidence for humans.

## 4.4 What the inductive biases of LMs can tell us about language

Since neural language models do have inductive biases, perhaps one reason for their success in learning language is that those inductive biases meaningfully align with linguistic structure. Indeed, a number of recent studies have investigated this possibility.

In one such study, Kallini et al. (2024) refute the claim that neural LMs can learn any language, including unnatural ones, equally well (Bolhuis et al., 2024). They compare the learning curves for the GPT-2 architecture trained on language modeling on English text against models trained on various transformations of the English text, designed to create languages that are intuitively 'impossible', but which still have the same level of overall predictability as the original English text. For example, one transformation applies a deterministic shuffling function to the tokens of English text, creating extraordinarily complex but deterministic word order rules that violate all known formal characterizations of syntax, and another transformation creates a new agreement marker that must appear exactly 4 tokens away from a verb, also an unnatural pattern. Kallini et al. (2024) find that the model learns from real English text consistently faster than these baselines (see also Mitchell and Bowers, 2020; Yang et al., 2025; Xu et al., 2025; Ziv et al., 2025, among others).

These results and others show that Transformers and related models have inductive biases that align with human language. However, the major determinant of inductive biases in LM is not that they are restricted to a particular formal language class, as might be expected from the generative linguistics paradigm. In fact, in terms of formal expressivity, it seems that Transformers are mismatched with the usual formal language classes used to characterize language. Whereas human language is sometimes characterized using (extensions of) the Chomsky–Schützenberger hierarchy (Chomsky and Schützenberger, 1963; Vijay-Shanker et al., 1987; Weir, 1988), which encompasses well-known classes such as regular and context-free languages, modern Transformers as they are currently applied (and other recently successful language model architectures such as State Space Models: Gu and Dao, 2024) seem to inhabit formal language classes defined by *circuit complexity* (Merrill et al., 2022; Strobl et al., 2024; Merrill et al., 2024), a formal language hierarchy which is orthogonal to the Chomsky–Schützenberger hierarchy.[13] To the extent that Transformers have inductive biases that help (or hinder) their language learning, they likely arise from something other than the expressive limits of their architecture.

[13] Interestingly, circuit complexity has also been used to characterize the computational capacity of biologically realistic populations of neurons (Maass, 1997; Maass and Markram, 2004). So human performance may also be ultimately limited in this way.

Below we consider two apparent learning biases of modern LMs which may be aligned with the structure of human language: information locality and low sensitivity.

**Information locality** Human languages are structured in a way such that elements that statistically predict each other are usually close to each other. For example, in phrase such as *big brown box*, the noun *box* and the adjective *brown* are highly predictive of each other—boxes, especially cardboard ones, are often brown, for many reasons—and so these words are likely to be close to each other: the alternate order *brown big box* sounds odd or like it is conveying some other special meaning (Futrell, 2019; Culbertson et al., 2020; Scontras, 2023; Dyer et al., 2023). Locality ideas of this kind pervade human language (Behaghel, 1930; Givón, 1991; Futrell, 2019; Mansfield, 2021; Hahn et al., 2021a; Mansfield and Kemp, 2023): words are usually contiguous units, usually modified by prefixes and suffixes (directly adjacent to them, ordered by 'relevance' to the root: Bybee, 1985; Saldana et al., 2024), and words linked by syntactic dependencies tend to be close to each other (Gibson, 1991, 1998; Liu, 2008; Liu et al., 2017), more than would expected under random grammars within a linguistically realistic formalism (Gildea and Temperley, 2007; Park and Levy, 2009; Gildea and Temperley, 2010; Futrell et al., 2015, 2020b).

Autoregressive LMs such as GPT-2 also show a bias towards information locality, as demonstrated by several of the experiments of Kallini et al. (2024) (discussed above): many of the counterfactual languages which are harder to learn in those experiments are also those that disrupt information locality. The bias towards locality seems to come from the next-token prediction task performed by LMs. The bias toward information locality is an 'ember of autoregression' in the terminology of McCoy et al. (2023), one which helps language learning and is likely shared with humans.

**Relatively low sensitivity** Another related inductive bias in Transformers is the bias toward learning functions with low sensitivity or low polynomial degree (Hahn et al., 2021b; Abbe et al., 2023; Bhattamishra et al., 2023). Sensitive functions are functions on input strings whose outputs change drastically based on small changes to the input. For example, a function on input bitstrings that counts the parity of the input—returning odd or even depending on the number of 1's in the input—is maximally sensitive and high-degree, because any single change to the input bits will flip the output of the function (O'Donnell, 2014). Although the Transformer architecture has the ability to represent highly sensitive functions in terms of its representational capacity, this turns out not to be relevant to its inductive bias in learning. Rather, the bias toward low-sensitivity functions comes from the shape of the loss landscape induced by the model. In particular, any parameter setting representing a highly sensitive function in the Transformer architecture must be *brittle*, meaning that a small change to the parameters would make the Transformer produce some different, lower-sensitivity function (Hahn and Rofin, 2024). Thus high-sensitivity functions are unlikely to be reached through a gradient-descent-based learning process.

Human languages, viewed (for example) as functions from strings to meanings or to grammaticality judgments, also seem to be *relatively* low-sensitivity (Hahn et al., 2021b). We do not find human languages where, for example, a word is well-formed if and only if it contains an odd number of some certain segment, even though it would be easy to express this language using a regular grammar.

However, like information locality, low sensitivity is not an absolute formal restriction on languages. For example, calculating the meaning of iterated negation is like a parity function: What does it mean when someone says they are *not not not not not not* wearing a tie? These high-sensitivity parts of language are nonetheless rare in usage, and difficult to understand in practice. Relative low sensitivity is an important statistical property of human language, perhaps reflecting a general cognitive constraint for humans, which has come to light as the result of taking the inductive biases of LM architectures seriously.

## 4.5 Convergence of LMs and psycholinguistics on predictive processing

It is not a coincidence that the basic paradigm for training language models—incremental prediction of upcoming input—mirrors ideas about how human language processing works. In early neural network work, Elman (1990b) cites two reasons for choosing to use a prediction task in training his models: "minimal assumptions about special knowledge required for training" and the fact that "much of what listeners do involves anticipation of future input". Indeed, there is a deep functional convergence between the predictive approach to language modeling and the ways that human language is processed and structured.

Elman's statement about what listeners do is not speculative. Experimental psycholinguistics has delivered a picture of human language processing that is deeply entwined with the task of autoregressive prediction, mirroring other areas of cognition (Rao and Ballard, 1999; Bar, 2009; Friston and Kiebel, 2009; Ryskin and Nieuwland, 2023). In comprehension (the process of decoding meaning from linguistic form in real time), the way that human processing works is influenced by the predictability of each word (or sub-word unit) conditional on previous words (Hale, 2001; Levy, 2008; Wilcox et al., 2020, 2023b; Xu et al., 2023). Furthermore, language comprehension is highly *incremental* (Tanenhaus et al., 1995; Smith and Levy, 2013), meaning that information of different kinds is processed and integrated as quickly as possible as it comes in. There is also a high degree of incrementality in language production (Ferreira and Swets, 2002), and incremental predictability plays a key role there too: for example, disfluencies and pauses in speech happen before unpredictable words (Goldman-Eisler, 1957; Beattie and Butterworth, 1979; Dammalapati et al., 2019; Futrell, 2023), and unpredictable words are enunciated more slowly (Bell et al., 2009; Upadhye and Futrell, 2025).

Many of these studies about the effects of predictability in language processing would have been impossible without large language models. Even after converging on the idea that prediction is an important part of language processing, psycholinguistics as a field was hampered by the fact that strong probabilistic models for language were not available, resulting in conclusions based on what now seem to be fairly weak models and methods of estimating incremental probability (Frank and Bod, 2011; Fossum and Levy, 2012; Smith and Levy, 2013). Now, neural models make it possible to probe the limits of predictive models of language processing (Wilcox et al., 2020, 2023b; Xu et al., 2023), including the discovery that certain phenomena seem to go beyond the simplest predictive models (van Schijndel and Linzen, 2018, 2021; Huang et al., 2024; Staub, 2024), and that human processing is better characterized by prediction under resource constraints similar to but more severe than those present in neural models (Futrell et al., 2020a; Hahn et al., 2022; Kuribayashi et al., 2022; Timkey and Linzen, 2023; Oh and Schuler, 2023; De Varda and Marelli, 2024; Clark et al., 2025).

## 4.6 Functional explanations for human language

LMs also orient us toward another route to explanatory adequacy in linguistic theory, one which posits that the form of language is related to its function: communication of thought and social coordination under general cognitive constraints on how language is produced and comprehended (Chomsky, 2005; Gibson et al., 2019; Levshina, 2022; Bickel et al., 2024). The *functionalist* school of linguistics, which is often contrasted to the generative or formalist approach (Newmeyer, 1998), holds that the structure of language ultimately reflects constraints and pressures arising from the way language is actually used (Hawkins, 1994, 2004, 2014; Haspelmath, 2008; Comrie, 1989)—perhaps motivated by factors such as a language's "niche" such that languages with larger communities of speakers or more second language learners might have different pressures (Lupyan and Dale, 2010; Raviv et al., 2019).

We believe that linguistics and language modeling can make (and in fact already have made) fruitful contact at this functional level of explanation. By analogy, bird wings and airplane wings are very different things, but they share the *function* of flying in the Earth's atmosphere, and so they are both shaped by the constraints of aerodynamics (Marr, 1982; Gill, 1995). Similarly, neural language models and human language processing mechanisms are different things, but at the level of function, they both encode and decode information in linguistic strings *incrementally* and *predictively*. To the extent that human language is structured in a way that supports this kind of incremental prediction (for example, through information locality), it is also aligned with the strengths and inductive biases of language models.

Indeed, language models have already proved a key tool in functionalist models of why languages are the way they are. A sizable literature exists about what kinds of languages emerge in simulated populations of agents who communicate by encoding and decoding meanings into strings using various neural architectures, and what constraints (on the environment, the agents, or the task) are necessary for the emergent languages to resemble natural language (Lazaridou et al., 2017; Mordatch and Abbeel, 2018; Steinert-Threlkeld, 2020; Kuciński et al., 2021; Chaabouni et al., 2021). For example, Hahn et al. (2020) show that certain universal properties of word order (Greenberg, 1963; Dryer, 1992) can be derived by finding grammars that optimize (1) the ease of recovering a parse tree from a string, and (2) the ease of incrementally predicting each word, with both factors operationalized using neural networks (see also Kuribayashi et al., 2024). Clark et al. (2023) show that natural language word order seems to be structured in a way that minimizes the *variance* of word-by-word surprisals, again measured using neural LMs, in keeping with the theory of Uniform Information Density (Fenk and Fenk, 1980; Levy and Jaeger, 2007; Jaeger, 2010; Jaeger and Tily, 2011). More generally, because human language processing is highly probabilistic and predictive, theories of linguistic structure based on functional constraints must be evaluated using a strong probabilistic predictive model. Language models provide exactly that.

## 4.7 Upshots for linguistics beyond language structure

In much of this piece, we have focused on the upshot of LMs for linguistic structure, particularly morphosyntax. We did so for two main reasons: first, because of the centrality of these topics within linguistics over the last 60+ years; and, second, because mastery of linguistic *form* emerged in models earlier than other high-level abilities (Mahowald et al., 2024). Taken together, these developments have led to important insights about human language learning and processing.

But LMs have also made recent striking gains in domains like reasoning, logic, and long-form dialog. These abilities seem to emerge not just from pretraining, but from techniques like supervised fine-tuning, instruction-tuning, and reinforcement learning from human feedback whereby models are given specific feedback to make them more aligned with human behavior on specific tasks (Ouyang et al., 2022; Bai et al., 2022; Achiam et al., 2023). The importance of inference-time compute—whereby models learn in-context and perform additional computation during the process of generating text—has also become apparent for tasks like math and reasoning (OpenAI et al., 2024; Marjanović et al., 2025).

As such, just as it is fruitful to explore what inductive biases and representational mechanisms underlie the emergence of syntax in LMs, it is increasingly fruitful to study what inductive biases and structures are necessary and/or sufficient for other abilities in LMs like reasoning (Marjanović et al., 2025), theory of mind (Hu et al., 2025), planning (Liu et al., 2023), and other aspects of higher-level cognition. Many of our same arguments hold in these domains as well: the upshot of the modern view of learning, the Contravariance Principle, the role of the statistical tradition, real patterns.

Moreover, as LMs continue to develop, they increasingly are becoming part of our speech communities. Millions of humans regularly interact with chatbots, in some cases carrying on long extended dialogs with chatbots playing various roles. And, increasingly, text read by humans is written or co-written by AI. How will the presence of AI change language? How should we think about dialogs between human and non-human interlocutors? What social consequences will emerge from this new paradigm? Linguistics as a field, particularly sociocultural linguistics and related disciplines, is well-positioned to be at the forefront of answering these questions (Bucholtz and Hall, 2005; Meyerhoff, 2006).

## 5 Conclusion

As Norm and Claudette would agree, this is a remarkable and pivotal moment in linguistics. Even ten years ago, it was not obvious that Claudette's dream of models that produce fluent and coherent text would be possible in her lifetime—or ever. It was also not obvious that, if it were to happen, it would happen using statistical systems trained largely on next-word prediction, using massive amounts of data. These systems don't use Norm's hand-crafted rules. They don't rely on insights from generative linguistic theory, although they do draw on decades of linguistic work in distributional semantics and statistical language learning.

Rather than resisting this development, we contend it offers a golden opportunity for linguistics by enabling new kinds of research and opening up vistas of new hypotheses, methods, and research questions. Some of these questions will be new, like what kinds of artificial neural architectures best capture language, what kinds of biases they hold, and what the sources of those biases are. But we also have highlighted ways in which these methods can make progress on some of the oldest and most venerable questions in linguistics: what must be true of the input data for certain structures to be learned in principle, what are the constraints on what languages are possible, how does the form of language relate to its function.

In response to Piantadosi's claim that language models refute Chomsky, Kodner et al. (2023) wrote a piece skeptical of the role of LMs in linguistic theory called "Why linguistics will thrive in the 21st century". We would go a step further and argue that, with the 21st century a quarter over, linguistics *is* thriving. But we think it's doing so not in spite of language models, but in part because of them and the opportunities they open up for new research directions and the light they shed on old questions.

Therefore, we reject the dichotomy in the discourse around Piantadosi's (2023) paper, which often seems to pit language models against linguistics. Linguistic structure, as described in linguistic theories, is real and important even if language models learn those structures emergently in a complex statistical way. Conversely, language models can point the way to new ways of thinking about the learning, processing, and fundamental nature of language.

But we do think there are some major revisions to some earlier accepted dogma that are warranted, including:

- Much of the field of linguistics has assumed that the form of language must be explained in terms of a symbolic formalism that constrains the forms of possible languages, and that this grammar formalism must represent innate human constraints on language, and that these constraints are logically necessary for language to be learned. The success of machine learning models should orient us away from this paradigm, because it shows that this approach is not necessary to characterize learning, and it opens up a new universe of statistical, quantitative, and functional theories to constrain the forms of possible languages and explain why only humans have language, while at the same time providing the tools to test those theories.

- Relatedly, the idea that the formal structure of linguistic competence should be the main focus of theoretical linguistics, which precedes other forms of study, is no longer tenable—language models reveal and allow us to characterize new dimensions of language such as how corpora of text express world knowledge, how the structure of usage supports learning or doesn't, and how fundamental information-processing constraints shape the way that language is represented in brains and machines.

- The success of machine learning models shows us how language can be represented in ways that are graded, probabilistic, and fuzzy, which should move us away from an insistence on discrete categorical frameworks. They also reveal potential soft constraints on the structure of language.

- The ontological basis for linguistic structure need not lie in an innate genetic endowment. Linguistic structure is a real pattern, just as real and worthy of study whether it lies innate in the human genome or whether it is learned entirely through inductive statistical learning with domain-general biases. Linguistic phenomena do not become less interesting when they are learnable in this way, they become *more* interesting.

More broadly, language models show that linguistics needs the rest of science. We will not answer the most fundamental questions about language by focusing *only* on language, using *only* methods that are tailor-made for linguistics. Just as neural networks revealed principles that are now fundamental in our understanding of vision, they are now in a position to revolutionize our understanding of language. Just as physics was revolutionized by thermodynamic ideas which arose from the ad-hoc analysis of complex machines, linguistics stands to benefit from new ways of thinking about computation arising from the analysis of neural networks. There is now massive intellectual activity spurred by language models in cognitive science, philosophy, physics, statistical learning theory, and information theory. The science of language can draw ideas, methods, and inspiration from all of this by maintaining a spirit of deep, curious, open-minded engagement and integration. Human language is in many ways unique as a natural phenomenon, and this means that the science of language should integrate with other fields of science *more*, in order to find deeper unities and to delineate and explain how and why language is special.

In particular, we are optimistic about significantly revising what Stabler (1983) said in doubting that "any close connections between linguistic theory and biology will be forthcoming". He continues: "it is a crucial advantage of the computational approach that it has a functionalist vocabulary that does not require a type reduction to physicalistic concepts [...] since this seems to be required in linguistics and in psychological accounts of language processing and visual perception". We agree that there is a crucial advantage to the functionalist vocabulary used in linguistic theory. But by analogy to recent work in vision, we are optimistic about the role that artificial neural models will play in bridging the gap between theory and biology.

This progress will come through an expansive linguistics — expansive in the breadth and diversity of languages it considers, expansive in its methods, and expansive in its connections to related fields. Baroni (2022) lamented that, as of a 2021 exploration of citation records, Linzen et al.'s seminal work on subject-verb agreement was largely uncited within the linguistics community. But, since then, there are increasingly researchers using language models to ask questions that are informed by and which can inform linguistic theory.

On the computer science side, while there is a persistent stereotype that the Natural Language Processing (NLP) community cares mostly about engineering, we have found there to be significant interest in fundamentally linguistic questions. Of the 7 papers that won Best Paper Awards at ACL 2024 (the largest annual conference focused on NLP), 1 was a direct response to a claim by Chomsky about language learning (Kallini et al., 2024), 1 was about inductive biases in models of relevance to questions about constraints on language (Hahn and Rofin, 2024), 1 was a theoretically motivated paper studying satisfiability in natural language of relevance to old questions about the complexity of language (Madusanka et al., 2024), 1 was a new method for measuring memorization in models and is of relevance for studying trade-offs in memorization vs. generalization in natural language (Lesci et al., 2024), 2 were about reconstructing or recovering ancient languages (Lu et al., 2024; Guan et al., 2024), and 1 presented an open-access multilingual model for broadening coverage of under-resourced languages (Üstün et al., 2024). Every one of these papers has relevance to scientific questions about human language, and in most cases one or more of the authors works in a linguistics department and/or has a degree in linguistics.

Thus, we don't see ourselves as calling for radical change, but rather as acknowledging the interdisciplinary work that is already thriving in the 21st century. Linguistically informed computational work is increasingly taking place within linguistics departments, where computational researchers are working alongside syntacticians, semanticists, phonologists, language documentation experts, sociocultural linguists, and experts in a wide variety of languages and language families. We think this is a very good development—one that both Norm and Claudette should embrace.

## 6 Acknowledgments

We thank Chris Potts for pointing out that he independently used a similar title in a talk called "Linguists for Deep Learning; or: How I Learned to Stop Worrying and Love Neural Networks" and encouraging us to use our title anyway. We thank David Beaver, Joan Bresnan, Ted Gibson, Greg Hickok, Laura Kalin, Andy Kehler, Robbie Kubala, Harvey Lederman, Connor Mayer, Kanishka Misra, Andrea Moro, Lisa Pearl, Jon Rawski, Greg Scontras, Alex Warstadt, Steve Wechsler, Xin Xie, and three anonymous reviewers for helpful comments.

## 7 Financial Support / Funding statement

KM was supported by NSF CRII grant 2104995.

## 8 Competing Interest statement

KM has done contract work for OpenAI, unrelated to the content of this manuscript.